\documentclass[sigconf]{acmart}

\usepackage{bm}
\usepackage{algorithmic}
\usepackage{graphicx}
\usepackage{textcomp}
\usepackage{xcolor}
\usepackage{array} 
\usepackage{tabularx}
\usepackage{subcaption}
\usepackage{booktabs}
\usepackage{multirow}
\usepackage{float}
\usepackage[normalem]{ulem}
\useunder{\uline}{\ul}{}

\begin{document}
\title{A Multi-scale Representation Learning Framework for Long-Term Time Series Forecasting}

\author{Boshi Gao, Qingjian Ni\textsuperscript{*}, Fanbo Ju, Yu Chen, Ziqi Zhao}
\affiliation{
  \institution{Southeast University}
  \city{Nanjing}
  \country{China}
}
\email{gaobs@seu.edu.cn, nqj@seu.edu.cn, 213234010@seu.edu.cn, yu_chen@seu.edu.cn, ziqizhao@seu.edu.cn}

\thanks{\textsuperscript{*}Corresponding author}

\renewcommand\footnotetextcopyrightpermission[1]{}
\settopmatter{printacmref=false}
\acmConference[]{Preprint}{May 2025}

\begin{abstract}
Long-term time series forecasting (LTSF) offers broad utility in practical settings like energy consumption and weather prediction. Accurately predicting long-term changes, however, is demanding due to the intricate temporal patterns and inherent multi-scale variations within time series. This work confronts key issues in LTSF, including the suboptimal use of multi-granularity information, the neglect of channel-specific attributes, and the unique nature of trend and seasonal components, by introducing a proficient MLP-based forecasting framework. Our method adeptly disentangles complex temporal dynamics using clear, concurrent predictions across various scales. These multi-scale forecasts are then skillfully integrated through a system that dynamically assigns importance to information from different granularities, sensitive to individual channel characteristics. To manage the specific features of temporal patterns, a two-pronged structure is utilized to model trend and seasonal elements independently. Experimental results on eight LTSF benchmarks demonstrate that MDMixer improves average MAE performance by 4.64\% compared to the recent state-of-the-art MLP-based method (TimeMixer), while achieving an effective balance between training efficiency and model interpretability.
\end{abstract}


\maketitle

\section{Introduction}
Long-term time series forecasting (LTSF) has been widely applied in various fields, such as energy consumption prediction~\cite{martin2010}, climate forecasting~\cite{zheng2015}, and traffic flow prediction~\cite{yin2021}, aiming to predict future time series based on historical data through modeling. Deep learning, with its powerful representational capabilities, has gained increasing attention in time series forecasting. Various deep learning architectures have been designed with clever strategies to capture temporal variations, such as models based on RNN~\cite{LSTNet}, CNN~\cite{MICN, TimesNet, ModernTCN}, Transformer~\cite{Informer, Autoformer, Fedformer, PatchTST, iTransformer}, and MLP~\cite{DLinear, RLinear, NHITS}. 

However, real-world time series data often exhibit intricate temporal patterns and multi-granularity dependencies, where multiple periodic variations (short-term, medium-term, and long-term) and trend patterns may occur simultaneously. For instance, electricity consumption data can exhibit sharp short-term fluctuations at an hourly resolution reflecting daily activity patterns, while monthly aggregated data may reveal broader seasonal cycles and long-term demand trends. This entanglement of multi-granularity information is one of the core challenges faced by current LTSF models~\cite{Kim2025}. Our central hypothesis is that if these diverse temporal components can be effectively disentangled and strategically recombined, it is possible to substantially enhance both the accuracy and robustness of forecasting performance.

In recent years, Transformer-based models have been increasingly applied to LTSF and have demonstrated strong modeling capabilities for time series data~\cite{Informer, Autoformer, Fedformer}. However, studies have shown that the permutation-invariance of the self-attention mechanism can lead to the loss of temporal information~\cite{DLinear}, and a simpler linear model can outperform almost all previous Transformer-based models~\cite{DLinear,TiDE}. While the attention mechanism aims to capture global dependencies, it often lacks explicit disentanglement of multi-granularity information. This can lead to performance degradation, potentially due to overfitting~\cite{tslanet}, and also results in a substantial decrease in training efficiency. Alternatively, simple linear models are efficient and can capture specific periodic patterns in time series~\cite{RLinear}; however, they struggle to capture intricate patterns due to the inherent simplicity of Linear-based models~\cite{MoLE}. In contrast, MDMixer aims to strike a new balance between efficiency and representational capacity, global and local processing, and multi-scale information handling.

\begin{figure}[t]
  \centering
  \includegraphics[width=0.9\columnwidth]{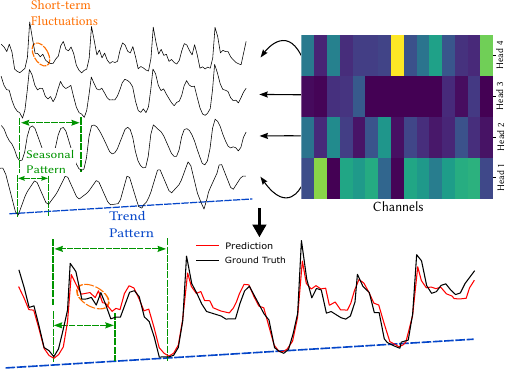}
  \caption{The core concept of MDMixer: (Top-Left) Parallel prediction of dynamic patterns at different temporal granularities using MPP; (Top-Right) Adaptive, channel-specific fusion of multi-granularity information, with weights determined by AMWG; (Bottom) Generation of long-term forecasts that closely track the ground truth. We observe that coarse-grained predictions capture overall seasonal and trend patterns, while fine-grained predictions extract short-term fluctuations.}
  \label{mixing}
  \vspace{-1em}
\end{figure}

Motivated by the above observations, we propose MDMixer, a novel architecture aimed at resolving the challenges associated with processing multi-scale temporal information. Technically, to explicitly disentangle patterns across multiple temporal granularities and to learn more expressive representations, we propose Multi-granularity Parallel Predictor (MPP) block, which applies multiple prediction heads to the input sequence for multi-granularity prediction. Additionally, to intelligently and hierarchically fuse information across different temporal granularities, we introduce the Multi-granularity Iterative Mixing (MIM) block. More importantly, a key observation in multivariate time series is that each channel often exhibits characteristic patterns across different temporal scales. Consequently, the simple aggregation or fixed weight assignment used by previous methods~\cite{timemixer,wpmixer} is insufficient to dynamically integrate such multi-granularity information. To address this limitation, we propose Adaptive Multi-granularity Weighting Gate (AMWG), a channel-dependent mechanism designed to achieve more refined information fusion by adaptively allocating weights across both granularity and channel dimensions, while also promoting information interaction between channels and enhancing the model's interpretability. Figure~\ref{mixing} presents the intermediate representations and prediction results produced by MDMixer on the Electricity dataset and illustrates the core concept of our method. The figure illustrates that the model's coarse-grained predictions can extract the smooth long-term periodicities and overall trends within the sequence. This coarse-grained information is crucial for understanding the macroscopic direction of the sequence. Concurrently, the model's fine-grained predictions capture the rapid fluctuations and specific details occurring within these broader coarse-grained cycles. Such fine-grained information is vital for accurately forecasting precise values and identifying turning points in the short term. Beyond disentangling and fusing information across various temporal granularities, MDMixer further refines its predictions by acknowledging the fundamental differences between seasonal and trend patterns. Specifically, since seasonal components exhibit periodic fluctuations, linear models effectively capture these variations, while MLP models excel at capturing the nonlinear dynamics of trends~\cite{RLinear}. We employ a dual-branch structure with Linear-based and MLP-based predictors to handle these components separately, enabling our model to fully exploit the strengths of each model and enhance the ability to capture different temporal patterns.

Our contributions are summarized as follows:

\begin{itemize}
  \item We propose a novel \textbf{Multi-granularity Predictor and Mixer} module that captures temporal representations at varying granularities and progressively fuses them in a hierarchical manner, to effectively capture both short-term fluctuations and long-term dependencies.

  \item We design a \textbf{Adaptive Multi-granularity Weighting Gate} that adaptively assigns fusion weights across both temporal granularities and variable channels. In addition, we introduce a \textbf{granularity-aligned supervision strategy} that aligns intermediate predictions with multi-scale targets to guide representation learning.

  \item We employ a \textbf{trend-seasonal dual-branch architecture} that uses linear layers for seasonal components and MLPs for trend components. This design not only improves the expressiveness of the model but can also be readily integrated into other Linear-based forecasting models to enhance their performance.

  \item Extensive experiments on multiple long-term time series forecasting benchmarks demonstrate that our proposed model, \textbf{MDMixer}, achieves state-of-the-art (SOTA) performance with superior computational efficiency and enhanced model interpretability.
\end{itemize}

\section{Related Work}

\subsection{MLP-based Time Series Forecasting Models}
MLP-based models have been increasingly applied in LTSF in recent years due to their efficiency and strong performance~\cite{DLinear, RLinear, NBeats, NHITS, FITS, TSMixer}. As a representative work of MLP-based models, N-BEATS~\cite{NBeats} makes interpretable forecasting based on backward and
forward residual links and a deep stack of fully connected layers. Recent research by DLinear~\cite{DLinear} challenges the effectiveness of transformers in time series forecasting. The model performs series decomposition before applying linear regression and outperforms all previously proposed Transformer-based models. RLinear~\cite{RLinear} observes that linear mapping can effectively capture periodic features in time series and combines reversible normalization to improve overall forecasting performance. 
As an MLP-based model, MDMixer further enhances MLP's performance in LTSF through leveraging multi-granularity processing and adaptive mixing.
\subsection{Decomposition of Time Series}
As one of the most common decomposition methods in time series analyses, seasonal-trend decomposition~\cite{decomp1,decomp2} separates the raw series into trend, seasonal, cyclical, and residual parts, making it easier to predict~\cite{tssurvey}. It can be achieved by using filters or exponential smoothing~\cite{decomp3}. Autoformer~\cite{Autoformer} firstly introduces the idea of decomposition to deep models and proposes a series decomposition block as a basic module to extract the seasonal and trend parts of input series, which has been widely used in the following works~\cite{preformer, tempo}. MICN~\cite{MICN} uses multiple pooling filters and takes the average of them as the final seasonal and trend series. MDMixer further leverages the advantages of the decomposition architecture, adopting a dual-branch structure to model the trend and seasonal components using different modules, based on their distinct characteristics.
\subsection{Mixing Networks}
Mixing is an effective method for integrating information, which improves the model's representational capacity and performance by combining different types of information. TSMixer~\cite{TSMixer} effectively extracts information across channels by introducing mixing operations along the feature dimension. TimeMixer~\cite{timemixer} proposes a multi-scale mixing architecture built upon MLP to integrate the diverse pattern information manifested in time series at different sampling scales. Unlike TimeMixer, which generates multiscale series via input downsampling and employs a simpler ensemble for future predictions, our MDMixer directly generates multi-granularity predictions from the input sequence using parallel heads, which enhances input information fidelity. And we further introduce an Adaptive Multi-granularity Weighting Gate for a more sophisticated, channel-aware fusion. Scaleformer~\cite{scaleformer} proposes a general multi-scale framework that iteratively refines a forecasted time series at multiple scales with shared weights. SOFTS~\cite{softs} utilizes its centralized STAR module for channel interaction by fusing a global core. In this paper, we further investigate the mixing architectures in LSTF. MDMixer captures multi-level patterns in time series through adaptive mixing of predictions at multiple granularities.

\section{Proposed Method}
\begin{figure*}[htbp]
    \centering
    \includegraphics[width=0.95 \textwidth]{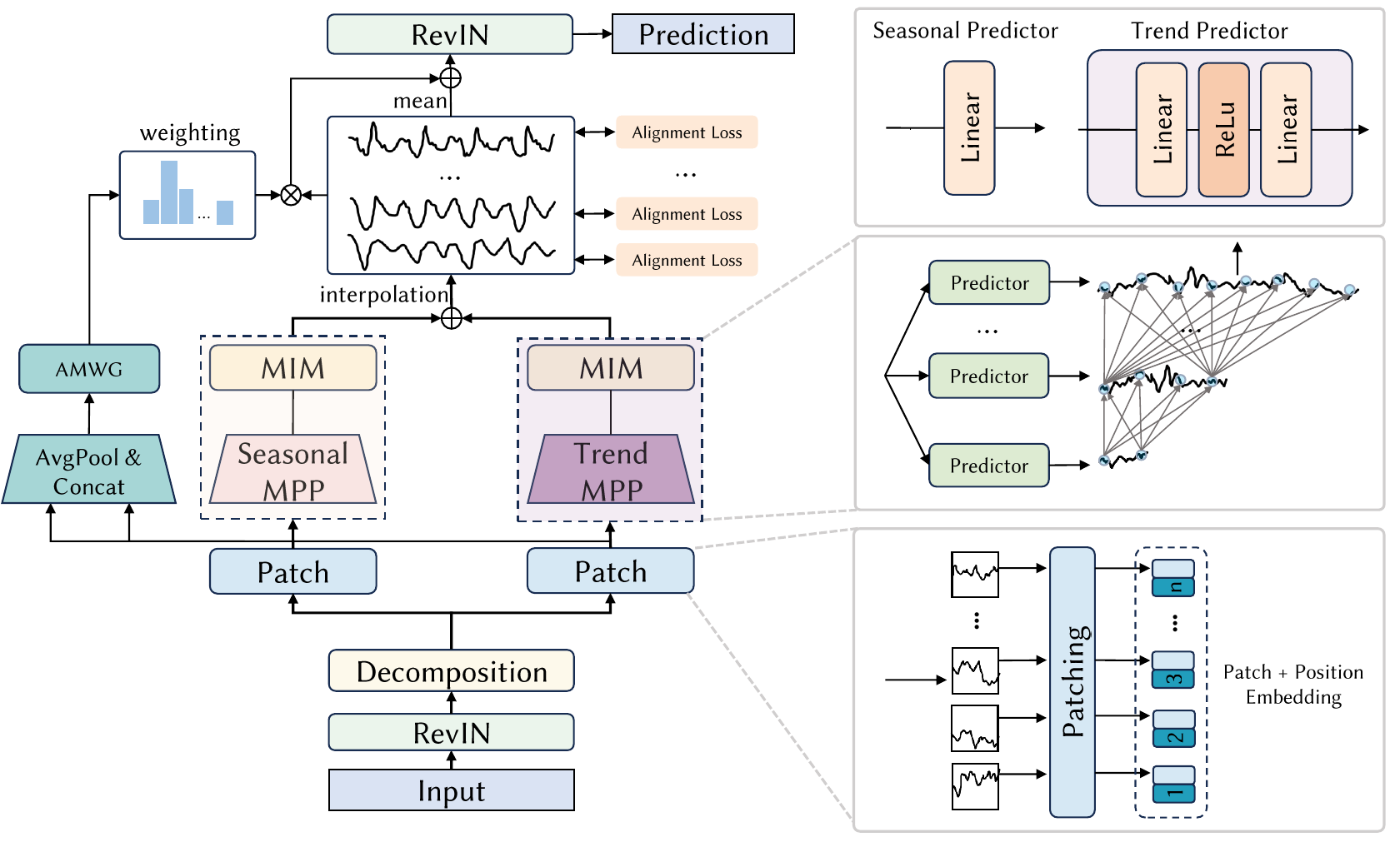} 
    \caption{Overview of MDMixer architecture. Multivariate time series are decomposed into trend and seasonal components. The respective branches process them using MLP-based and Linear-based modules for prediction. The Multi-granularity Parallel Predictor (MPP) and the Multi-granularity Iterative Mixer (MIM) are responsible for multi-granularity prediction and fusion, respectively. Predictions at the same granularity level from both the trend and seasonal branches are summed and aligned with the downsampled target sequence through the computation of alignment loss. The Adaptive Multi-granularity Weighting Gate (AMWG) takes the combined patch embeddings as input and produces dynamic, channel-specific weights to aggregate the multi-granularity predictions.}
    \label{fig: architecture}
\end{figure*}

In multivariate time series forecasting, given historical observations $\bm{X} = \{\bm{x}_1, \dots, \bm{x}_T\} \in \mathbb{R}^{T \times C}$ with $T$ time steps and $C$ variates, we predict the future $F$ time steps $\bm{Y^\ast} = \{\bm{x}_{T+1}, \dots, \bm{x}_{T+F}\} \in \mathbb{R}^{F \times C}$. The architecture of MDMixer is illustrated in Figure~\ref{fig: architecture}.
\subsection{Instance Normalization and Dual-branch Decomposition}

\subsubsection{Instance Normalization}
This technique was recently proposed to mitigate the distribution shift between training and testing data~\cite{RevIN, RevIN2}, a common challenge in real-world time series that can significantly degrade model performance if not addressed~\cite{cicd}. It has been widely adopted in various studies~\cite{PatchTST, patchmixer, RLinear}. The method involves normalizing each time series instance $x^{(i)}$ to have zero mean and unit standard deviation. Specifically, the normalization is applied to $x^{(i)}$ after input, and the original mean and standard deviation are restored to the output before generating predictions.

\subsubsection{Dual-branch Decomposition}
Seasonal-trend decomposition is a standard method in time series analysis to make raw data more predictable. Specifically, they use a moving average kernel on the input sequence to extract the trend and seasonal component $\bm{X}^{t}$ and $\bm{X}^{s}$. This process is expressed as follows:
\begin{equation}
\label{decomp}
\begin{aligned}
\bm{X}^{t}&=\text{AvgPool}(\text{Padding}(\bm{X})),\\
\bm{X^{s}}&=\bm{X}-\bm{X^{t}}.
\end{aligned}
\end{equation}

Recent research~\cite{RLinear} indicates that while linear models effectively capture time series seasonality, they exhibit poor performance when modeling trend components. This may be due to the accumulation of predictive errors for the trend component as timestamps increase or as the trend becomes more pronounced. Furthermore, studies have shown that real-world time series often exhibit complex nonlinear trends~\cite{nonlinearities,chen2025nonlinear}. MLP can outperform traditional linear regression models in capturing such complex nonlinear patterns, owing to its inherent nonlinear architecture and the theoretical foundation provided by the universal approximation theorem~\cite{multilayer}. Inspired by this, we propose a novel dual-branch architecture to model the seasonal and trend components, respectively. 

Technically, we designed two versions of the Multi-granularity Parallel Predictor (MPP) block, with their predictors based on Linear and MLP models, respectively. The Linear-based version is used to model seasonal components, while the MLP-based version is designed to model trend components. Through this design, we leverage the strength of the Linear model in capturing seasonal components while addressing its weaknesses in trend modeling.

\subsection{Patch Embedding}
Patch Embedding divides the input time series into smaller patches to capture local features and short-term dependencies. In this process, input sequence  \( \bm{X} \in \mathbb{R}^{T \times C} \) is divided into patches in the dimension of $T$, which may overlap or remain non-overlapping. With a patch length \( P \) and a stride \( S \), this results in a sequence of patches \( \bm{X}_p \in \mathbb{R}^{C \times N \times P} \), where \( N \) is the number of patches,
$N = \left\lfloor \frac{T - P}{S} \right\rfloor + 2$.
Zero padding is applied at the end of the sequence to ensure an appropriate length for patching. Each patch \( \bm{X}_p \) represents a segment of the original series. A linear embedding layer: $\mathbb{R}^{P} \rightarrow \mathbb{R}^{D} $ transforms the patches into a latent space of dimension $D$, with a learnable positional encoding \( {\bm{X}_{pos}} \in \mathbb{R}^{C \times N \times D} \) added to preserve temporal order:
\begin{equation}
\bm{X}_d = \text{Embedding}(\bm{X}_p)+\bm{X}_{pos} \in\mathbb{R}^{C \times N\times D},
\end{equation}
where \( \bm{X}_d \) represents the embedded series. This patch embedding step is applied separately to the trend and seasonal components $\bm{X^{s}}$ and $\bm{X^{t}}$, resulting in $\bm{X_d^{s}}$ and $\bm{X_d^{t}}$ $\in \mathbb{R}^{C \times N \times D}$.

\subsection{Multi-granularity Predictor and Mixer}

To capture the complex entanglement of temporal patterns across different scales in time series, we design a unified Multi-granularity Predictor and Mixer (MPM) block. This module aims to generate multiple levels of predictive representations that reflect short-term and long-term dependencies. The MPM block is composed of two main components: the Multi-granularity Parallel Predictor (MPP) and the Multi-granularity Iterative Mixer (MIM).

\subsubsection{Multi-granularity Parallel Predictor}

Our MPP block is designed to address multi-granularity dependencies in time series by uncovering the complex entanglement patterns within the data. By extracting features at multiple granularities, this approach enables the model to capture short-term fluctuations and long-term trends simultaneously. Each predictor head functions as an independent feature extractor, focusing on different scales, thus allowing the model to learn both local and global patterns effectively.

Given the embedded seasonal and trend components $\bm{X}_d^{s}$ and $\bm{X}_d^{t} \in \mathbb{R}^{C \times N \times D}$, we flatten the last two dimensions and obtain $\mathbb{R}^{C \times (N \cdot D)}$. The MPP applies parallel prediction heads to extract features at multiple temporal granularities by applying $H$ independent heads to $\bm{X}_d^{s}$ and $\bm{X}_d^{t}$ separately. For each head $i \in \{1, 2, \ldots, H\}$, we define the prediction length as: $G_i = g \cdot i$, where $g=\frac{F}{H}$ is the base granularity unit depending on $H$ and $G_i$ denotes the output length of the $i$-th head. It provides a systematic and progressive increase in predictive granularity across the heads. This ensures a comprehensive and balanced coverage of the temporal spectrum, from coarser-scale patterns (shorter $G_i$) to finer-scale fluctuations (longer $G_i$). Specifically, each head produces:

\begin{align}
\bm{Z}_{i}^{s} &= \mathcal{L}_{\text{linear}}^{(i)}(\bm{X}_d^{s}) \in \mathbb{R}^{C \times G_i}, \\
\bm{H}_{i}^{t} &= \text{ReLU}(\mathcal{L}_1^{(i)}(\bm{X}_d^{t})) \in \mathbb{R}^{C \times H_{\text{hid}}}, \label{eq:mlp1} \\
\bm{Z}_{i}^{t} &= \mathcal{L}_2^{(i)}(\bm{H}_{i}^{t}) \in \mathbb{R}^{C \times G_i},\label{eq:mlp2}
\end{align}
where $\mathcal{L}_{\text{linear}}^{(i)}: \mathbb{R}^{C \times (N \cdot D)} \rightarrow \mathbb{R}^{C \times G_i}$ maps the seasonal branch directly to the target granularity length;
$\mathcal{L}_1^{(i)}: \mathbb{R}^{C \times (N \cdot D)} \rightarrow \mathbb{R}^{C \times H_{\text{hid}}}$ projects the trend branch to a hidden space of width $H_{\text{hid}}$ with \text{ReLU} activation, and $\mathcal{L}_2^{(i)}: \mathbb{R}^{C \times H_{\text{hid}}} \rightarrow \mathbb{R}^{C \times G_i}$ maps the intermediate trend representation to the target output length $G_i$.

Each head extracts features independently and makes multi-granularity predictions, ensuring diversity in the learned representations. These outputs form two independent multi-granularity prediction sets:
\begin{equation}
\mathcal{Z}^{s} = \left\{ \bm{Z}_1^s, \bm{Z}_2^s, \ldots, \bm{Z}_H^s \right\}, \quad
\mathcal{Z}^{t} = \left\{ \bm{Z}_1^t, \bm{Z}_2^t, \ldots, \bm{Z}_H^t \right\}.
\end{equation}

\subsubsection{Multi-granularity Iterative Mixer}
To combine the predictions generated by MPP, the MIM block progressively fuses the outputs using an iterative layer-by-layer accumulation in a coarse-to-fine manner. This coarse-to-fine strategy builds on our earlier observation: overall patterns (like trends and seasonality) from coarser scales first establish a contextual base. Finer scales then enrich this base with short-term fluctuation details. MIM allows the distinct features from MPP at each level of granularity to be enriched through interaction with other scales, ensuring that scale-specific information is preserved while also benefiting from broader contextual insights.

Specifically, we apply the MIM separately on the seasonal and trend prediction sets $\mathcal{Z}^{s}$ and $\mathcal{Z}^{t}$. We initialize $\bm{Y}_1^s = \bm{Z}_1^s$ and $\bm{Y}_1^t = \bm{Z}_1^t$. The mixing is performed in an iterative manner for $i$ from $2$ to $H$:

\begin{equation}
\quad \bm{Y}_i^s = \bm{Z}_i^s + \mathcal{M}_i^s( \bm{Y}_{i-1}^s) \in \mathbb{R}^{C \times G_i},
\end{equation}
\begin{equation}
\quad \bm{Y}_i^t = \bm{Z}_i^t + \mathcal{M}_i^t( \bm{Y}_{i-1}^t) \in \mathbb{R}^{C \times G_i},
\end{equation}
\begin{equation}
\bm{Y}_i = \bm{Y}_i^s + \bm{Y}_i^t \in \mathbb{R}^{C \times G_i},
\end{equation}
where $\mathcal{M}_i^s$ and $\mathcal{M}_i^t$ denote the linear layer at the $i$-th granularity level. Each mapping operates as $\mathcal{M}_i^s : \mathbb{R}^{C \times G_{i-1}} \rightarrow \mathbb{R}^{C \times G_i}$. These mappings integrate the current prediction $\bm{Z}_i$ with the cumulative context $\bm{Y}_{i-1}$ from the previous head, producing refined outputs. The final prediction $\bm{Y}_i$ at $i$-th granularity is obtained by summing the seasonal and trend outputs.

\subsection{Adaptive Multi-granularity Weighting Gate}
In the above MPM block, the prediction heads are designed to capture dynamic features at distinct temporal scales. We notice that naively assigning a scalar weight to each head for fusion fails to account for the varying sensitivity of different variables to different granularities. To this end, we propose an Adaptive Multi-granularity Weighting Gate (AMWG) fusion mechanism, which simultaneously models fusion weights along both the granularity and the variable (channel) dimensions. This mechanism constructs a contextual representation for fusion by globally aggregating the embedded seasonal and trend components, and employs a gating network to generate a weight tensor of size $H \times C$.

Specifically, to gain a comprehensive representation of the overall characteristics of each channel across both its seasonal and trend components, we apply global average pool on $\bm{X}_d^{s}$ and $\bm{X}_d^{t}$ over their last two dimensions (temporal and feature dimensions, resulting in a flattened dimension of $N \cdot D$) and concatenate them to form the gate input:
\begin{align}
\bar{\bm{X}}^{s} &= \text{AvgPool}(\bm{X}_d^{s}) \in \mathbb{R}^{C}, \\
\bar{\bm{X}}^{t} &= \text{AvgPool}(\bm{X}_d^{t}) \in \mathbb{R}^{C}, \\
\bm{G}_{\text{in}} &= \text{Concat}(\bar{\bm{X}}^{s}, \bar{\bm{X}}^{t}) \in \mathbb{R}^{2C}.
\end{align}
This step constructs a core representation by aggregating global channel information, which helps to mitigate noise arising from individual anomalous fluctuations and enhances the robustness of the model. Moreover, our concatenation strategy preserves the separation of trend and seasonal components, enabling the downstream network to capture more flexible nonlinear mappings.

Then, a two-layer feedforward network with a ReLU activation is employed as the gating network, providing capacity to learn the complex mapping from the concatenated context $\bm{G}_{\text{in}}$ to the channel-wise granularity weights $\bm{W}'$:
\begin{equation}
\bm{W}' = \phi\left(\mathcal{G}_2(\text{ReLU}(\mathcal{G}_1(\bm{G}_{\text{in}})))\right) \in \mathbb{R}^{H \times C},
\end{equation}
where $\mathcal{G}_1$ and $\mathcal{G}_2$ are linear layers, \( \mathcal{G}_1: \mathbb{R}^{2C} \rightarrow \mathbb{R}^{H_{\text{hid}}} \), \( \mathcal{G}_2: \mathbb{R}^{H_{\text{hid}}} \rightarrow \mathbb{R}^{H \cdot C} \), and \( \phi(\cdot) \) denotes the reshape operation: \( H \cdot C \rightarrow  H \times C \).

For each channel \(c\), the fusion weights over heads are normalized via:
\begin{equation}
\bm{W}_{[:,c]} = \text{Softmax}(\bm{W}^\prime_{[:,c]}),
\end{equation}
where softmax is applied over the head dimension \(H\).

After computing the weights $\bm{W}$, the next step is to integrate the predictions from the $H$ prediction heads. Since each prediction $\bm{Y}_i \in \mathbb{R}^{C \times G_i}$ has a different length $G_i$, all outputs are upsampled to the target length $F$ to produce the final unified prediction through linear interpolation:
\begin{equation}
\widetilde{\bm{Y}}_i = \text{Interpolation}(\bm{Y_i}) \in \mathbb{R}^{C \times F}.
\end{equation}

The gating weights are broadcast to match the prediction dimension and applied as element-wise multipliers. And the weighted multi-granularity results are added residually to the averaged outputs of all prediction heads:
\begin{equation}
\bm{Y} = \sum_{i=1}^{H} \bm{W}_{[i, :]} \otimes \widetilde{\bm{Y}}_i + \frac{1}{H} \sum_{i=1}^{H} \widetilde{\bm{Y}}_i \in \mathbb{R}^{C \times F},
\end{equation}
where \( \otimes \) denotes broadcasting multiplication over the last dimension $F$. The averaged unweighted predictions are included as a stable baseline, enhancing overall prediction robustness. By combining the weighted predictions of different granularities across different channels, we obtain the final prediction $\bm{Y}$.

It is important to highlight that AMWG adopts a channel-dependent fusion strategy. By leveraging a gating network driven by global channel-level context $\bm{G}_{\text{in}}$ to produce channel-wise weights $ \bm{W} \in  H \times C $, AMWG dynamically adjusts the importance of predictions at each granularity for every variable, thereby enabling more precise and context-aware information fusion.

\subsection{Loss Function}

To enhance the model's capacity for learning representations across different granularities, we adopt a two-part loss to supervise both the final output and the intermediate multi-granularity predictions.
\paragraph{Main Prediction Loss}
The primary loss is the L1 norm between ground truth sequence $\bm{Y}^{\ast}$ and model prediction $\bm{Y}$:
\begin{equation}
\mathcal{L}_{\text{main}} = \| \bm{Y} - \bm{Y}^{\ast} \|_1.
\end{equation}

\paragraph{Multi-granularity Alignment Loss}
To ensure that each head \( i \in \{1, \ldots, H\} \) captures the temporal structure at its respective granularity \( G_i \), we supervise its intermediate prediction \( \bm{Y_i} \in \mathbb{R}^{C \times G_i} \) using a down-sampled version of the ground truth \( \bm{Y}^{\ast} \). We construct the target sequences \( \bm{Y_i}^{\ast} \in \mathbb{R}^{C \times G_i} \) by applying average pooling to \( \bm{Y}^{\ast} \). The alignment loss for head \( i \) is computed as:
\begin{equation}
\mathcal{L}_\text{align}^{i} = \| \bm{Y_i} - \bm{Y_i}^{\ast} \|_1.
\end{equation}

\paragraph{Total Loss}
The total loss is defined as a combination of the main loss and the average alignment loss across all granularities:
\begin{equation}
\mathcal{L}_{\text{total}} =
\mathcal{L}_{\text{main}} + \alpha \cdot \frac{1}{H} \sum_{i=1}^{H} \mathcal{L}_\text{align}^{i}, 
\end{equation}
where \( \alpha \) is a weighting hyperparameter.

\section{Experiments}
\subsection{Forecasting Results}
\subsubsection{Datasets}
To evaluate the effectiveness of our proposed MDMixer, we conduct extensive experiments on eight widely used real-world datasets spanning various domains, including energy, economics, weather, and traffic. These datasets include: ETT~\cite{Informer}, Exchange~\cite{Autoformer}, Weather~\cite{Autoformer}, Electricity~\cite{Autoformer}, and Traffic~\cite{Autoformer}. We follow the standard protocol~\cite{Informer} and split all datasets into training, validation, and test sets in chronological order by the ratio of 6:2:2 for the ETT datasets and 7:1:2 for the other datasets. Table~\ref{datasets_summary} summarizes the details of datasets.
\begin{table}[H]
\caption{The details of datasets.}
\small
\renewcommand{\arraystretch}{0.8}
\centering
\resizebox{0.75\columnwidth}{!}{
\begin{tabular}{@{}cccc@{}}
\toprule
Datasets    & Channels & Frequency & Timesteps \\ \midrule
ETTh1       & 7        & 1 hour    & 17,420    \\
ETTh2       & 7        & 1 hour    & 17,420    \\
ETTm1       & 7        & 15 mins   & 69,680    \\
ETTm2       & 7        & 15 mins   & 69,680    \\
Exchange    & 8        & 1 day     & 7,588     \\
Weather     & 21       & 10 mins   & 52,696    \\
Electricity & 321      & 1 hour    & 26,304   \\
Traffic &  862   &   1 hour   &   17,544
\\\bottomrule

\end{tabular}
\label{datasets_summary}}
\end{table}

\begin{table*}[]
\centering
\caption{Multivariate long-term time series forecasting results. The input length $T=96$ for all baselines and the forecast horizon $F \in \{96,192,336,720\}$ is set for all datasets. The best results are highlighted in \textbf{bold} and the second best are \underline{underlined}.}
\label{LTSF result}
\small
\renewcommand{\arraystretch}{0.85}
\resizebox{\textwidth}{!}{%
\begin{tabular}{@{}ccllllllllllllllll@{}}
\toprule
\multicolumn{2}{c|}{Models}                                                                                     & \multicolumn{2}{c}{\begin{tabular}[c]{@{}c@{}}MDMixer\\ (ours)\end{tabular}} & \multicolumn{2}{c}{\begin{tabular}[c]{@{}c@{}}TimeMixer\\ (2024)\end{tabular}} & \multicolumn{2}{c}{\begin{tabular}[c]{@{}c@{}}iTransformer\\ (2024)\end{tabular}} & \multicolumn{2}{c}{\begin{tabular}[c]{@{}c@{}}RLinear\\ (2023)\end{tabular}} & \multicolumn{2}{c}{\begin{tabular}[c]{@{}c@{}}PatchTST\\ (2023)\end{tabular}} & \multicolumn{2}{c}{\begin{tabular}[c]{@{}c@{}}TimesNet\\ (2023)\end{tabular}} & \multicolumn{2}{c}{\begin{tabular}[c]{@{}c@{}}DLinear\\ (2023)\end{tabular}} & \multicolumn{2}{c}{\begin{tabular}[c]{@{}c@{}}MICN\\ (2023)\end{tabular}} \\ \midrule
\multicolumn{2}{c|}{Metric}                                                                                     & \multicolumn{1}{c}{MSE}               & \multicolumn{1}{c}{MAE}              & \multicolumn{1}{c}{MSE}                & \multicolumn{1}{c}{MAE}               & \multicolumn{1}{c}{MSE}                 & \multicolumn{1}{c}{MAE}                 & \multicolumn{1}{c}{MSE}               & \multicolumn{1}{c}{MAE}              & \multicolumn{1}{c}{MSE}               & \multicolumn{1}{c}{MAE}               & \multicolumn{1}{c}{MSE}               & \multicolumn{1}{c}{MAE}               & \multicolumn{1}{c}{MSE}               & \multicolumn{1}{c}{MAE}              & \multicolumn{1}{c}{MSE}             & \multicolumn{1}{c}{MAE}             \\ \midrule
\multicolumn{1}{c|}{\multirow{5}{*}{\rotatebox{90}{ETTh1}}}       & \multicolumn{1}{c|}{96}  & {\ul 0.379}                           & \textbf{0.386}                       & \textbf{0.375}                         & 0.400                                 & 0.386                                   & 0.405                                   & 0.386                                 & {\ul 0.395}                          & 0.414                                 & 0.419                                 & 0.384                                 & 0.402                                 & 0.386                                 & 0.400                                & 0.426                               & 0.446                               \\
\multicolumn{1}{c|}{}                                                                & \multicolumn{1}{c|}{192} & {\ul 0.434}                           & \textbf{0.418}                       & \textbf{0.429}                         & {\ul 0.421}                           & 0.441                                   & 0.436                                   & 0.437                                 & 0.424                                & 0.460                                 & 0.445                                 & 0.436                                 & 0.429                                 & 0.437                                 & 0.432                                & 0.454                               & 0.464                               \\
\multicolumn{1}{c|}{}                                                                & \multicolumn{1}{c|}{336} & \textbf{0.479}                        & \textbf{0.437}                       & 0.484                                  & 0.458                                 & 0.487                                   & 0.458                                   & \textbf{0.479}                        & {\ul 0.446}                          & 0.501                                 & 0.466                                 & 0.491                                 & 0.469                                 & {\ul 0.481}                           & 0.459                                & 0.493                               & 0.487                               \\
\multicolumn{1}{c|}{}                                                                & \multicolumn{1}{c|}{720} & \textbf{0.478}                        & \textbf{0.452}                       & 0.498                                  & 0.482                                 & 0.503                                   & 0.491                                   & {\ul 0.481}                           & {\ul 0.470}                          & 0.500                                 & 0.488                                 & 0.521                                 & 0.500                                 & 0.519                                 & 0.516                                & 0.526                               & 0.526                               \\ \cmidrule(l){2-18} 
\multicolumn{1}{c|}{}                                                                & \multicolumn{1}{c|}{Avg} & \textbf{0.442}                        & \textbf{0.423}                       & 0.447                                  & 0.440                                 & 0.454                                   & 0.448                                   & {\ul 0.446}                           & {\ul 0.434}                          & 0.469                                 & 0.455                                 & 0.458                                 & 0.450                                 & 0.456                                 & 0.452                                & 0.475                               & 0.480                               \\ \midrule
\multicolumn{1}{c|}{\multirow{5}{*}{\rotatebox{90}{ETTh2}}}       & \multicolumn{1}{c|}{96}  & \textbf{0.281}                        & \textbf{0.329}                       & 0.289                                  & 0.341                                 & 0.297                                   & 0.349                                   & {\ul 0.288}                           & {\ul 0.338}                          & 0.302                                 & 0.348                                 & 0.340                                 & 0.374                                 & 0.333                                 & 0.387                                & 0.372                               & 0.424                               \\
\multicolumn{1}{c|}{}                                                                & \multicolumn{1}{c|}{192} & \textbf{0.352}                        & \textbf{0.379}                       & {\ul 0.372}                            & 0.392                                 & 0.380                                   & 0.400                                   & 0.374                                 & {\ul 0.390}                          & 0.388                                 & 0.400                                 & 0.402                                 & 0.414                                 & 0.477                                 & 0.476                                & 0.492                               & 0.492                               \\
\multicolumn{1}{c|}{}                                                                & \multicolumn{1}{c|}{336} & {\ul 0.405}                           & {\ul 0.416}                          & \textbf{0.386}                         & \textbf{0.414}                        & 0.428                                   & 0.432                                   & 0.415                                 & 0.426                                & 0.426                                 & 0.433                                 & 0.452                                 & 0.452                                 & 0.594                                 & 0.541                                & 0.607                               & 0.555                               \\
\multicolumn{1}{c|}{}                                                                & \multicolumn{1}{c|}{720} & \textbf{0.411}                        & \textbf{0.431}                       & {\ul 0.412}                            & {\ul 0.434}                           & 0.427                                   & 0.445                                   & 0.420                                 & 0.440                                & 0.431                                 & 0.446                                 & 0.462                                 & 0.468                                 & 0.831                                 & 0.657                                & 0.824                               & 0.655                               \\ \cmidrule(l){2-18} 
\multicolumn{1}{c|}{}                                                                & \multicolumn{1}{c|}{Avg} & \textbf{0.362}                        & \textbf{0.389}                       & {\ul 0.364}                            & {\ul 0.395}                           & 0.383                                   & 0.407                                   & 0.374                                 & 0.399                                & 0.387                                 & 0.407                                 & 0.414                                 & 0.427                                 & 0.559                                 & 0.515                                & 0.574                               & 0.531                               \\ \midrule
\multicolumn{1}{c|}{\multirow{5}{*}{\rotatebox{90}{ETTm1}}}       & \multicolumn{1}{c|}{96}  & \textbf{0.309}                        & \textbf{0.339}                       & {\ul 0.320}                            & {\ul 0.357}                           & 0.334                                   & 0.368                                   & 0.355                                 & 0.376                                & 0.329                                 & 0.367                                 & 0.338                                 & 0.375                                 & 0.345                                 & 0.372                                & 0.365                               & 0.387                               \\
\multicolumn{1}{c|}{}                                                                & \multicolumn{1}{c|}{192} & \textbf{0.361}                        & \textbf{0.365}                       & \textbf{0.361}                         & {\ul 0.381}                           & 0.377                                   & 0.391                                   & 0.391                                 & 0.392                                & {\ul 0.367}                           & 0.385                                 & 0.374                                 & 0.387                                 & 0.380                                 & 0.389                                & 0.403                               & 0.408                               \\
\multicolumn{1}{c|}{}                                                                & \multicolumn{1}{c|}{336} & {\ul 0.397}                           & \textbf{0.390}                       & \textbf{0.390}                         & {\ul 0.404}                           & 0.426                                   & 0.420                                   & 0.424                                 & 0.415                                & 0.399                                 & 0.410                                 & 0.410                                 & 0.411                                 & 0.413                                 & 0.413                                & 0.436                               & 0.431                               \\
\multicolumn{1}{c|}{}                                                                & \multicolumn{1}{c|}{720} & {\ul 0.463}                           & \textbf{0.429}                       & \textbf{0.454}                         & 0.441                                 & 0.491                                   & 0.459                                   & 0.487                                 & 0.450                                & \textbf{0.454}                        & {\ul 0.439}                           & 0.478                                 & 0.450                                 & 0.474                                 & 0.453                                & 0.489                               & 0.462                               \\ \cmidrule(l){2-18} 
\multicolumn{1}{c|}{}                                                                & \multicolumn{1}{c|}{Avg} & {\ul 0.383}                           & \textbf{0.381}                       & \textbf{0.381}                         & {\ul 0.395}                           & 0.407                                   & 0.410                                   & 0.414                                 & 0.408                                & 0.387                                 & 0.400                                 & 0.400                                 & 0.406                                 & 0.403                                 & 0.407                                & 0.423                               & 0.422                               \\ \midrule
\multicolumn{1}{c|}{\multirow{5}{*}{\rotatebox{90}{ETTm2}}}       & \multicolumn{1}{c|}{96}  & \textbf{0.171}                        & \textbf{0.248}                       & {\ul 0.175}                            & {\ul 0.258}                           & 0.180                                   & 0.264                                   & 0.182                                 & 0.265                                & {\ul 0.175}                           & 0.259                                 & 0.187                                 & 0.267                                 & 0.193                                 & 0.292                                & 0.197                               & 0.296                               \\
\multicolumn{1}{c|}{}                                                                & \multicolumn{1}{c|}{192} & \textbf{0.235}                        & \textbf{0.292}                       & {\ul 0.237}                            & {\ul 0.299}                           & 0.250                                   & 0.309                                   & 0.246                                 & 0.304                                & 0.241                                 & 0.302                                 & 0.249                                 & 0.309                                 & 0.284                                 & 0.362                                & 0.284                               & 0.361                               \\
\multicolumn{1}{c|}{}                                                                & \multicolumn{1}{c|}{336} & \textbf{0.293}                        & \textbf{0.329}                       & {\ul 0.298}                            & {\ul 0.340}                           & 0.311                                   & 0.348                                   & 0.307                                 & 0.342                                & 0.305                                 & 0.343                                 & 0.321                                 & 0.351                                 & 0.369                                 & 0.427                                & 0.381                               & 0.429                               \\
\multicolumn{1}{c|}{}                                                                & \multicolumn{1}{c|}{720} & {\ul 0.393}                           & \textbf{0.388}                       & \textbf{0.391}                         & {\ul 0.396}                           & 0.412                                   & 0.407                                   & 0.407                                 & 0.398                                & 0.402                                 & 0.400                                 & 0.408                                 & 0.403                                 & 0.554                                 & 0.522                                & 0.549                               & 0.522                               \\ \cmidrule(l){2-18} 
\multicolumn{1}{c|}{}                                                                & \multicolumn{1}{c|}{Avg} & \textbf{0.273}                        & \textbf{0.314}                       & {\ul 0.275}                            & {\ul 0.323}                           & 0.288                                   & 0.332                                   & 0.286                                 & 0.327                                & 0.281                                 & 0.326                                 & 0.291                                 & 0.333                                 & 0.350                                 & 0.401                                & 0.353                               & 0.402                               \\ \midrule
\multicolumn{1}{c|}{\multirow{5}{*}{\rotatebox{90}{Exchange}}}    & \multicolumn{1}{c|}{96}  & \textbf{0.082}                        & \textbf{0.198}                       & 0.090                                  & 0.235                                 & {\ul 0.086}                             & 0.206                                   & 0.093                                 & 0.217                                & 0.088                                 & {\ul 0.205}                           & 0.107                                 & 0.234                                 & 0.088                                 & 0.218                                & 0.148                               & 0.278                               \\
\multicolumn{1}{c|}{}                                                                & \multicolumn{1}{c|}{192} & \textbf{0.173}                        & \textbf{0.294}                       & 0.187                                  & 0.343                                 & 0.177                                   & {\ul 0.299}                             & 0.184                                 & 0.307                                & {\ul 0.176}                           & {\ul 0.299}                           & 0.226                                 & 0.344                                 & {\ul 0.176}                           & 0.315                                & 0.271                               & 0.315                               \\
\multicolumn{1}{c|}{}                                                                & \multicolumn{1}{c|}{336} & 0.321                                 & {\ul 0.409}                          & 0.353                                  & 0.473                                 & 0.331                                   & 0.417                                   & 0.351                                 & 0.432                                & \textbf{0.301}                        & \textbf{0.397}                        & 0.367                                 & 0.448                                 & {\ul 0.313}                           & 0.427                                & 0.460                               & 0.427                               \\
\multicolumn{1}{c|}{}                                                                & \multicolumn{1}{c|}{720} & \textbf{0.835}                        & \textbf{0.687}                       & 0.934                                  & 0.761                                 & 0.847                                   & {\ul 0.691}                             & 0.886                                 & 0.714                                & 0.901                                 & 0.714                                 & 0.964                                 & 0.746                                 & {\ul 0.839}                           & 0.695                                & 1.195                               & 0.695                               \\ \cmidrule(l){2-18} 
\multicolumn{1}{c|}{}                                                                & \multicolumn{1}{c|}{Avg} & \textbf{0.353}                        & \textbf{0.397}                       & 0.391                                  & 0.453                                 & 0.360                                   & {\ul 0.403}                             & 0.379                                 & 0.418                                & 0.367                                 & 0.404                                 & 0.416                                 & 0.443                                 & {\ul 0.354}                           & 0.414                                & 0.519                               & 0.429                               \\ \midrule
\multicolumn{1}{c|}{\multirow{5}{*}{\rotatebox{90}{Weather}}}     & \multicolumn{1}{c|}{96}  & \textbf{0.152}                        & \textbf{0.191}                       & {\ul 0.163}                            & {\ul 0.209}                           & 0.174                                   & 0.214                                   & 0.192                                 & 0.232                                & 0.177                                 & 0.218                                 & 0.172                                 & 0.220                                 & 0.196                                 & 0.255                                & 0.198                               & 0.261                               \\
\multicolumn{1}{c|}{}                                                                & \multicolumn{1}{c|}{192} & \textbf{0.204}                        & \textbf{0.238}                       & {\ul 0.208}                            & {\ul 0.250}                           & 0.221                                   & 0.254                                   & 0.240                                 & 0.271                                & 0.225                                 & 0.259                                 & 0.219                                 & 0.261                                 & 0.237                                 & 0.296                                & 0.239                               & 0.299                               \\
\multicolumn{1}{c|}{}                                                                & \multicolumn{1}{c|}{336} & {\ul 0.260}                           & \textbf{0.280}                       & \textbf{0.251}                         & {\ul 0.287}                           & 0.278                                   & 0.296                                   & 0.292                                 & 0.307                                & 0.278                                 & 0.297                                 & 0.280                                 & 0.306                                 & 0.283                                 & 0.335                                & 0.285                               & 0.336                               \\
\multicolumn{1}{c|}{}                                                                & \multicolumn{1}{c|}{720} & {\ul 0.343}                           & \textbf{0.335}                       & \textbf{0.339}                         & {\ul 0.341}                           & 0.358                                   & 0.347                                   & 0.364                                 & 0.353                                & 0.354                                 & 0.348                                 & 0.365                                 & 0.359                                 & 0.345                                 & 0.381                                & 0.351                               & 0.388                               \\ \cmidrule(l){2-18} 
\multicolumn{1}{c|}{}                                                                & \multicolumn{1}{c|}{Avg} & \textbf{0.240}                        & \textbf{0.261}                       & \textbf{0.240}                         & {\ul 0.271}                           & {\ul 0.258}                             & 0.278                                   & 0.272                                 & 0.291                                & 0.259                                 & 0.281                                 & 0.259                                 & 0.287                                 & 0.265                                 & 0.317                                & 0.268                               & 0.321                               \\ \midrule
\multicolumn{1}{c|}{\multirow{5}{*}{\rotatebox{90}{Electricity}}} & \multicolumn{1}{c|}{96}  & {\ul 0.152}                           & {\ul 0.242}                          & 0.153                                  & 0.247                                 & \textbf{0.148}                          & \textbf{0.240}                          & 0.201                                 & 0.281                                & 0.181                                 & 0.270                                 & 0.168                                 & 0.272                                 & 0.197                                 & 0.282                                & 0.180                               & 0.293                               \\
\multicolumn{1}{c|}{}                                                                & \multicolumn{1}{c|}{192} & 0.167                                 & \textbf{0.253}                       & {\ul 0.166}                            & {\ul 0.256}                           & \textbf{0.162}                          & \textbf{0.253}                          & 0.201                                 & 0.283                                & 0.188                                 & 0.274                                 & 0.184                                 & 0.289                                 & 0.196                                 & 0.285                                & 0.189                               & 0.302                               \\
\multicolumn{1}{c|}{}                                                                & \multicolumn{1}{c|}{336} & {\ul 0.183}                           & \textbf{0.268}                       & 0.185                                  & 0.277                                 & \textbf{0.178}                          & {\ul 0.269}                             & 0.215                                 & 0.298                                & 0.204                                 & 0.293                                 & 0.198                                 & 0.300                                 & 0.209                                 & 0.301                                & 0.198                               & 0.312                               \\
\multicolumn{1}{c|}{}                                                                & \multicolumn{1}{c|}{720} & \textbf{0.204}                        & \textbf{0.289}                       & 0.225                                  & {\ul 0.310}                           & 0.225                                   & 0.317                                   & 0.257                                 & 0.331                                & 0.246                                 & 0.324                                 & 0.220                                 & 0.320                                 & 0.245                                 & 0.333                                & {\ul 0.217}                         & 0.330                               \\ \cmidrule(l){2-18} 
\multicolumn{1}{c|}{}                                                                & \multicolumn{1}{c|}{Avg} & \textbf{0.177}                        & \textbf{0.263}                       & 0.182                                  & 0.272                                 & {\ul 0.178}                                   & {\ul 0.270}                                   & 0.219                                 & 0.298                                & 0.205                                 & 0.290                                 & 0.193                                 & 0.295                                 & 0.212                                 & 0.300                                & 0.196                               & 0.309                               \\ \midrule
\multicolumn{1}{c|}{\multirow{5}{*}{\rotatebox{90}{Traffic}}}     & \multicolumn{1}{c|}{96}  & {\ul 0.439}                           & {\ul 0.275}                          & \multicolumn{1}{c}{0.462}              & \multicolumn{1}{c}{0.285}             & \multicolumn{1}{c}{\textbf{0.395}}      & \multicolumn{1}{c}{\textbf{0.268}}      & \multicolumn{1}{c}{0.649}             & \multicolumn{1}{c}{0.389}            & \multicolumn{1}{c}{0.462}             & \multicolumn{1}{c}{0.295}             & \multicolumn{1}{c}{0.593}             & \multicolumn{1}{c}{0.321}             & \multicolumn{1}{c}{0.650}             & \multicolumn{1}{c}{0.396}            & 0.577                               & 0.350                               \\
\multicolumn{1}{c|}{}                                                                & \multicolumn{1}{c|}{192} & {\ul 0.447}                           & \textbf{0.276}                       & \multicolumn{1}{c}{0.473}              & \multicolumn{1}{c}{0.296}             & \multicolumn{1}{c}{\textbf{0.417}}      & \multicolumn{1}{c}{\textbf{0.276}}      & \multicolumn{1}{c}{0.601}             & \multicolumn{1}{c}{0.366}            & \multicolumn{1}{c}{0.466}             & \multicolumn{1}{c}{0.296}             & \multicolumn{1}{c}{0.617}             & \multicolumn{1}{c}{0.336}             & \multicolumn{1}{c}{0.598}             & \multicolumn{1}{c}{0.370}            & 0.589                               & 0.356                               \\
\multicolumn{1}{c|}{}                                                                & \multicolumn{1}{c|}{336} & {\ul 0.468}                           & {\ul 0.288}                          & \multicolumn{1}{c}{0.498}              & \multicolumn{1}{c}{0.296}             & \multicolumn{1}{c}{\textbf{0.433}}      & \multicolumn{1}{c}{\textbf{0.283}}      & \multicolumn{1}{c}{0.609}             & \multicolumn{1}{c}{0.369}            & \multicolumn{1}{c}{0.482}             & \multicolumn{1}{c}{0.304}             & \multicolumn{1}{c}{0.629}             & \multicolumn{1}{c}{0.336}             & \multicolumn{1}{c}{0.605}             & \multicolumn{1}{c}{0.373}            & 0.594                               & 0.358                               \\
\multicolumn{1}{c|}{}                                                                & \multicolumn{1}{c|}{720} & {\ul 0.504}                           & {\ul 0.304}                          & \multicolumn{1}{c}{0.506}              & \multicolumn{1}{c}{0.313}             & \multicolumn{1}{c}{\textbf{0.467}}      & \multicolumn{1}{c}{\textbf{0.302}}      & \multicolumn{1}{c}{0.647}             & \multicolumn{1}{c}{0.387}            & \multicolumn{1}{c}{0.514}             & \multicolumn{1}{c}{0.322}             & \multicolumn{1}{c}{0.640}             & \multicolumn{1}{c}{0.350}             & \multicolumn{1}{c}{0.645}             & \multicolumn{1}{c}{0.394}            & 0.613                               & 0.361                               \\ \cmidrule(l){2-18} 
\multicolumn{1}{c|}{}                                                                & \multicolumn{1}{c|}{Avg} & {\ul 0.465}                           & {\ul 0.286}                          & \multicolumn{1}{c}{0.484}              & \multicolumn{1}{c}{0.297}             & \multicolumn{1}{c}{\textbf{0.428}}      & \multicolumn{1}{c}{\textbf{0.282}}      & \multicolumn{1}{c}{0.626}             & \multicolumn{1}{c}{0.378}            & \multicolumn{1}{c}{0.481}             & \multicolumn{1}{c}{0.304}             & \multicolumn{1}{c}{0.620}             & \multicolumn{1}{c}{0.336}             & \multicolumn{1}{c}{0.625}             & \multicolumn{1}{c}{0.383}            & 0.593                               & 0.356                               \\ \midrule
\multicolumn{4}{c|}{Improvement}                                                                                                                                                               & 2.50\%                                 & 4.64\%                                & 2.21\%                                  & 4.10\%                                  & 10.64\%                               & 8.09\%                               & 4.97\%                                & 5.34\%                                & 11.67\%                               & 8.83\%                                & 16.41\%                               & 14.89\%                              & 20.76\%                             & 16.49\%                             \\ \bottomrule
\end{tabular}%
}
\end{table*}

\begin{table*}[htbp]
\centering
\caption{Comparison of the MSE and MAE results for our proposed dual-branch framework version (denoted "Dual-Branch") with respective baselines ($\ast$ denotes our modified baselines). The best results are shown in \textbf{Bold}. Our method outperforms the vanilla version on the vast majority of datasets.}
\label{dual-branch}
\resizebox{\textwidth}{!}{%
\tiny
\renewcommand{\arraystretch}{0.65}
\begin{tabular}{@{}ccllllllllllllll@{}}
\toprule
\multicolumn{2}{c|}{Models}                                                                   & \multicolumn{2}{c}{DLinear} & \multicolumn{2}{c|}{\begin{tabular}[c]{@{}c@{}}DLinear\\ (Dual-Branch)\end{tabular}} & \multicolumn{2}{c}{RLinear$\ast$} & \multicolumn{2}{c}{RMLP$\ast$}        & \multicolumn{2}{c|}{\begin{tabular}[c]{@{}c@{}}RLinear\\ (Dual-Branch)\end{tabular}} & \multicolumn{2}{c}{NLinear$\ast$}     & \multicolumn{2}{c}{\begin{tabular}[c]{@{}c@{}}NLinear\\ (Dual-Branch)\end{tabular}} \\ \midrule
\multicolumn{2}{c|}{Metric}                                                                  & MSE      & MAE              & MSE                            & \multicolumn{1}{l|}{MAE}                           & MSE      & MAE              & MSE            & MAE            & MSE                            & \multicolumn{1}{l|}{MAE}                           & MSE            & MAE            & MSE                                      & MAE                                      \\ \midrule
\multicolumn{1}{c|}{\multirow{5}{*}{\rotatebox{90}{ETTm1}}}       & \multicolumn{1}{c|}{96}  & 0.345    & 0.372            & \textbf{0.335}                 & \multicolumn{1}{l|}{\textbf{0.369}}                & 0.347    & 0.366            & 0.324          & 0.362          & \textbf{0.321}                 & \multicolumn{1}{l|}{\textbf{0.358}}                & 0.348          & 0.370          & \textbf{0.333}                           & \textbf{0.367}                           \\
\multicolumn{1}{c|}{}                                             & \multicolumn{1}{c|}{192} & 0.380    & \textbf{0.389}   & \textbf{0.371}                 & \multicolumn{1}{l|}{\textbf{0.389}}                & 0.386    & 0.385            & 0.368          & 0.384          & \textbf{0.364}                 & \multicolumn{1}{l|}{\textbf{0.382}}                & 0.388          & 0.389          & \textbf{0.372}                           & \textbf{0.388}                           \\
\multicolumn{1}{c|}{}                                             & \multicolumn{1}{c|}{336} & 0.413    & \textbf{0.413}   & \textbf{0.406}                 & \multicolumn{1}{l|}{0.416}                         & 0.418    & 0.405            & 0.397          & 0.406          & \textbf{0.395}                 & \multicolumn{1}{l|}{\textbf{0.403}}                & 0.420          & \textbf{0.411} & \textbf{0.405}                           & 0.415                                    \\
\multicolumn{1}{c|}{}                                             & \multicolumn{1}{c|}{720} & 0.474    & 0.453            & \textbf{0.467}                 & \multicolumn{1}{l|}{\textbf{0.448}}                & 0.478    & \textbf{0.439}   & 0.464          & 0.445          & \textbf{0.462}                 & \multicolumn{1}{l|}{0.442}                         & 0.482          & \textbf{0.446} & \textbf{0.465}                           & 0.447                                    \\ \cmidrule(l){2-16} 
\multicolumn{1}{c|}{}                                             & \multicolumn{1}{c|}{Avg} & 0.403    & 0.407            & \textbf{0.395}                 & \multicolumn{1}{l|}{\textbf{0.405}}                & 0.407    & 0.399            & 0.388          & 0.399          & \textbf{0.386}                 & \multicolumn{1}{l|}{\textbf{0.396}}                & 0.409          & \textbf{0.404} & \textbf{0.394}                           & \textbf{0.404}                           \\ \midrule
\multicolumn{1}{c|}{\multirow{5}{*}{\rotatebox{90}{ETTm2}}}       & \multicolumn{1}{c|}{96}  & 0.193    & 0.292            & \textbf{0.179}                 & \multicolumn{1}{l|}{\textbf{0.265}}                & 0.182    & 0.264            & 0.174          & 0.257          & \textbf{0.172}                 & \multicolumn{1}{l|}{\textbf{0.253}}                & \textbf{0.181} & \textbf{0.263} & 0.185                                    & 0.274                                    \\
\multicolumn{1}{c|}{}                                             & \multicolumn{1}{c|}{192} & 0.284    & 0.362            & \textbf{0.248}                 & \multicolumn{1}{l|}{\textbf{0.315}}                & 0.247    & 0.305            & \textbf{0.237} & 0.301          & 0.238                          & \multicolumn{1}{l|}{\textbf{0.297}}                & \textbf{0.245} & \textbf{0.303} & 0.249                                    & 0.321                                    \\
\multicolumn{1}{c|}{}                                             & \multicolumn{1}{c|}{336} & 0.369    & 0.427            & \textbf{0.324}                 & \multicolumn{1}{l|}{\textbf{0.367}}                & 0.308    & 0.343            & 0.296          & 0.338          & \textbf{0.295}                 & \multicolumn{1}{l|}{\textbf{0.334}}                & \textbf{0.306} & \textbf{0.341} & 0.322                                    & 0.365                                    \\
\multicolumn{1}{c|}{}                                             & \multicolumn{1}{c|}{720} & 0.554    & 0.522            & \textbf{0.468}                 & \multicolumn{1}{l|}{\textbf{0.469}}                & 0.408    & 0.399            & \textbf{0.390} & \textbf{0.392} & 0.391                          & \multicolumn{1}{l|}{0.393}                         & \textbf{0.406} & \textbf{0.397} & 0.457                                    & 0.461                                    \\ \cmidrule(l){2-16} 
\multicolumn{1}{c|}{}                                             & \multicolumn{1}{c|}{Avg} & 0.350    & 0.401            & \textbf{0.305}                 & \multicolumn{1}{l|}{\textbf{0.354}}                & 0.286    & 0.328            & \textbf{0.274} & 0.322          & \textbf{0.274}                 & \multicolumn{1}{l|}{\textbf{0.319}}                & \textbf{0.284} & \textbf{0.326} & 0.303                                    & 0.355                                    \\ \midrule
\multicolumn{1}{c|}{\multirow{5}{*}{\rotatebox{90}{Weather}}}     & \multicolumn{1}{c|}{96}  & 0.196    & 0.255            & \textbf{0.173}                 & \multicolumn{1}{l|}{\textbf{0.228}}                & 0.191    & 0.233            & 0.165          & 0.210          & \textbf{0.164}                 & \multicolumn{1}{l|}{\textbf{0.209}}                & 0.193          & 0.238          & \textbf{0.175}                           & \textbf{0.232}                           \\
\multicolumn{1}{c|}{}                                             & \multicolumn{1}{c|}{192} & 0.237    & 0.296            & \textbf{0.216}                 & \multicolumn{1}{l|}{\textbf{0.266}}                & 0.235    & 0.267            & \textbf{0.213} & \textbf{0.252} & \textbf{0.213}                 & \multicolumn{1}{l|}{0.253}                         & 0.240          & 0.271          & \textbf{0.217}                           & \textbf{0.270}                           \\
\multicolumn{1}{c|}{}                                             & \multicolumn{1}{c|}{336} & 0.283    & 0.335            & \textbf{0.266}                 & \multicolumn{1}{l|}{\textbf{0.307}}                & 0.287    & 0.303            & 0.269          & \textbf{0.292} & \textbf{0.268}                 & \multicolumn{1}{l|}{0.294}                         & 0.292          & 0.307          & \textbf{0.264}                           & \textbf{0.304}                           \\
\multicolumn{1}{c|}{}                                             & \multicolumn{1}{c|}{720} & 0.345    & 0.381            & \textbf{0.336}                 & \multicolumn{1}{l|}{\textbf{0.358}}                & 0.359    & 0.349            & \textbf{0.344} & 0.345          & 0.345                          & \multicolumn{1}{l|}{\textbf{0.343}}                & 0.366          & \textbf{0.355} & \textbf{0.333}                           & \textbf{0.355}                           \\ \cmidrule(l){2-16} 
\multicolumn{1}{c|}{}                                             & \multicolumn{1}{c|}{Avg} & 0.265    & 0.317            & \textbf{0.248}                 & \multicolumn{1}{l|}{\textbf{0.290}}                & 0.268    & 0.288            & 0.248          & \textbf{0.275} & \textbf{0.247}                 & \multicolumn{1}{l|}{\textbf{0.275}}                & 0.273          & 0.293          & \textbf{0.247}                           & \textbf{0.290}                           \\ \midrule
\multicolumn{1}{c|}{\multirow{5}{*}{\rotatebox{90}{Electricity}}} & \multicolumn{1}{c|}{96}  & 0.197    & 0.282            & \textbf{0.182}                 & \multicolumn{1}{l|}{\textbf{0.267}}                & 0.198    & 0.275            & 0.172          & 0.261          & \textbf{0.169}                 & \multicolumn{1}{l|}{\textbf{0.259}}                & 0.198          & 0.276          & \textbf{0.184}                           & \textbf{0.270}                           \\
\multicolumn{1}{c|}{}                                             & \multicolumn{1}{c|}{192} & 0.196    & 0.285            & \textbf{0.185}                 & \multicolumn{1}{l|}{\textbf{0.273}}                & 0.198    & 0.278            & 0.180          & 0.269          & \textbf{0.177}                 & \multicolumn{1}{l|}{\textbf{0.266}}                & 0.198          & 0.279          & \textbf{0.186}                           & \textbf{0.274}                           \\
\multicolumn{1}{c|}{}                                             & \multicolumn{1}{c|}{336} & 0.209    & 0.301            & \textbf{0.199}                 & \multicolumn{1}{l|}{\textbf{0.290}}                & 0.212    & 0.293            & 0.195          & \textbf{0.283} & \textbf{0.194}                 & \multicolumn{1}{l|}{\textbf{0.283}}                & 0.213          & 0.294          & \textbf{0.201}                           & \textbf{0.291}                           \\
\multicolumn{1}{c|}{}                                             & \multicolumn{1}{c|}{720} & 0.245    & 0.333            & \textbf{0.234}                 & \multicolumn{1}{l|}{\textbf{0.320}}                & 0.254    & 0.326            & 0.236          & 0.317          & \textbf{0.235}                 & \multicolumn{1}{l|}{\textbf{0.315}}                & 0.255          & 0.327          & \textbf{0.238}                           & \textbf{0.324}                           \\ \cmidrule(l){2-16} 
\multicolumn{1}{c|}{}                                             & \multicolumn{1}{c|}{Avg} & 0.212    & 0.300            & \textbf{0.200}                 & \multicolumn{1}{l|}{\textbf{0.288}}                & 0.215    & 0.293            & 0.196          & 0.283          & \textbf{0.194}                 & \multicolumn{1}{l|}{\textbf{0.281}}                & 0.216          & 0.294          & \textbf{0.202}                           & \textbf{0.290}                           \\ \bottomrule
\end{tabular}%
}
\end{table*}
\subsubsection{Baselines and Metrics}
As baselines, we select state-of-the-art and representative models in LTSF domain, including (1) Transformer-based methods: iTransformer~\cite{iTransformer}, PatchTST~\cite{PatchTST}; (2) MLP-based methods: TimeMixer~\cite{timemixer}, DLinear~\cite{DLinear}, RLinear~\cite{RLinear}; and (3) CNN-based methods: MICN~\cite{MICN}, TimesNet~\cite{TimesNet}. We evaluate model performance using Mean Squared Error (MSE) and Mean Absolute Error (MAE), as these metrics are standard for quantifying forecasting accuracy and robustness in time series prediction.
\subsubsection{Implementation Details}
Our training, validation, and test sets are zero-mean normalized with the mean and standard deviation of the training set in order to be consistent with prior work. We use PyTorch to implement all neural networks and train the model on an NVIDIA TITAN RTX GPU. The networks are trained by the AdamW optimizer with a learning rate of 1e-2 or 1e-3. We use MAE as the loss function. The look-back window size is fixed to 96 for all models, and the horizon varies from 96 to 720. The default configuration of MDMixer consists of a hidden size of 64, a patch length of 32, a stride length of 16, a head number of 8, and the alpha in the loss function is 0.01. For relatively large datasets (Electricity and Traffic), we set the hidden size to 128 and the head number to 16. To ensure robustness and minimize the impact of random fluctuations, we repeat each experiment 3 times with different seeds and report the average performance across all runs.
\subsubsection{LTSF Results}
The multivariate LTSF results are presented in Table~\ref{LTSF result}. The "Improvement" row indicates the average reduction in MSE and MAE compared to the baseline across all datasets. Overall, MDMixer consistently outperforms across most datasets. Quantitatively, compared to the strongest Transformer-based model (iTransformer), MDMixer reduces MSE and MAE by 2.21\% and 4.10\%, respectively. Furthermore, relative to the best-performing MLP-based model (TimeMixer), MDMixer achieves improvements of 2.50\% in MSE and 4.64\% in MAE. Notably, our approach surpasses PatchTST with a reduction of 4.97\% on MSE and 5.34\% on MAE, indicating that relying solely on the patch strategy, without multi-granularity interactions, does not yield optimal performance. MDMixer also brings significant performance improvements over RLinear and DLinear, indicating that simple linear models struggle to capture complex patterns in temporal data.
\subsection{Dual-Branch Augmentation for Linear Models}
To further validate the effectiveness and generalizability of our proposed dual-branch decomposition architecture, we conduct experiments by transplanting it into other linear models, including RLinear, DLinear, and NLinear~\cite{DLinear}. We also choose an MLP-based model, RMLP~\cite{RLinear}, for comparison. For clarity in Table~\ref{dual-branch}, models marked with an asterisk (e.g., RLinear$\ast$) represent our modified baseline versions. Models marked with "(Dual-Branch)" represent applying our proposed method. Specifically, for DLinear, which is originally a dual-branch architecture, we replace the trend component with a single-hidden-layer MLP from MDMixer, which is the same as Equations \ref{eq:mlp1} and \ref{eq:mlp2}, while keeping the rest of the structure unchanged. As for RLinear, RMLP, and NLinear, since their original structure does not include a decomposition module, we introduce a standard trend-seasonal decomposition, the same as Equation \ref{decomp}, as a baseline and then apply our proposed method for comparison. This excludes the performance improvement relative to the baseline brought by the trend-seasonal decomposition operation. 

Our experimental results are presented in Table~\ref{dual-branch}. The results show that our method brings consistent improvements to linear models, with an average improvement of 6.3\% for DLinear and 4.5\% for RLinear$\ast$. Compared to RMLP$\ast$, our method also achieves comparable performance. However, since RMLP$\ast$ employs an MLP to model the seasonal component, our method requires fewer parameters (50.2K vs. 34.8K in the Electricity dataset). We recommend that linear models employ linear and MLP modules to model the seasonal and trend components, respectively, thereby leveraging the strength of linear models in capturing seasonal patterns while improving the accuracy of trend forecasting.

\subsection{Ablation Study}
\begin{figure}[h]
    \centering
    \begin{subfigure}[b]{0.49\columnwidth}
        \centering
        \includegraphics[width=\columnwidth]{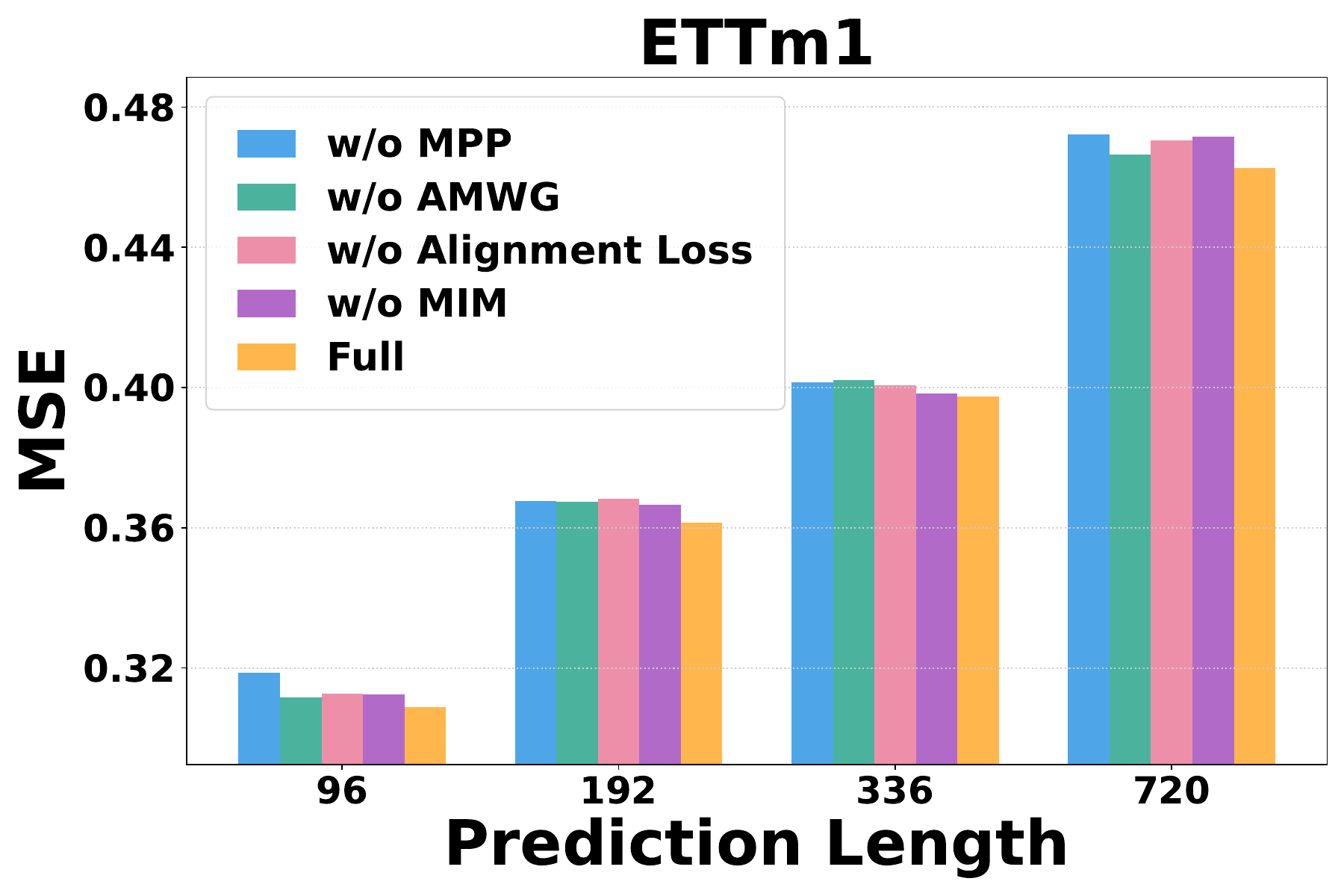}
        \label{fig:image1}
    \end{subfigure} \hfill
    \begin{subfigure}[b]{0.49\columnwidth}
        \centering
        \includegraphics[width=\columnwidth]{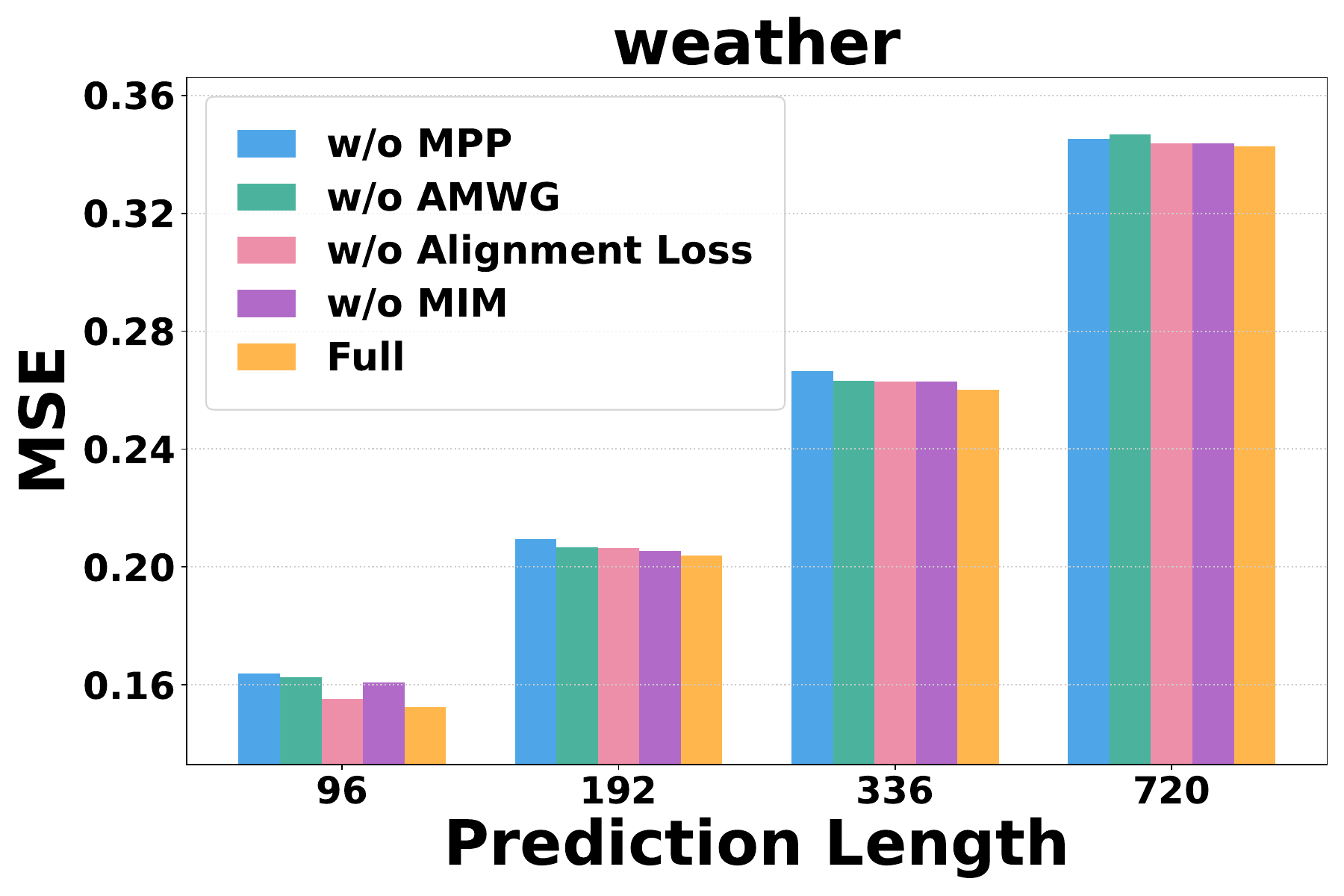}
        \label{fig:image2}
    \end{subfigure}

    \vspace{-1em}

    \begin{subfigure}[b]{0.49\columnwidth}
        \centering
        \includegraphics[width=\columnwidth]{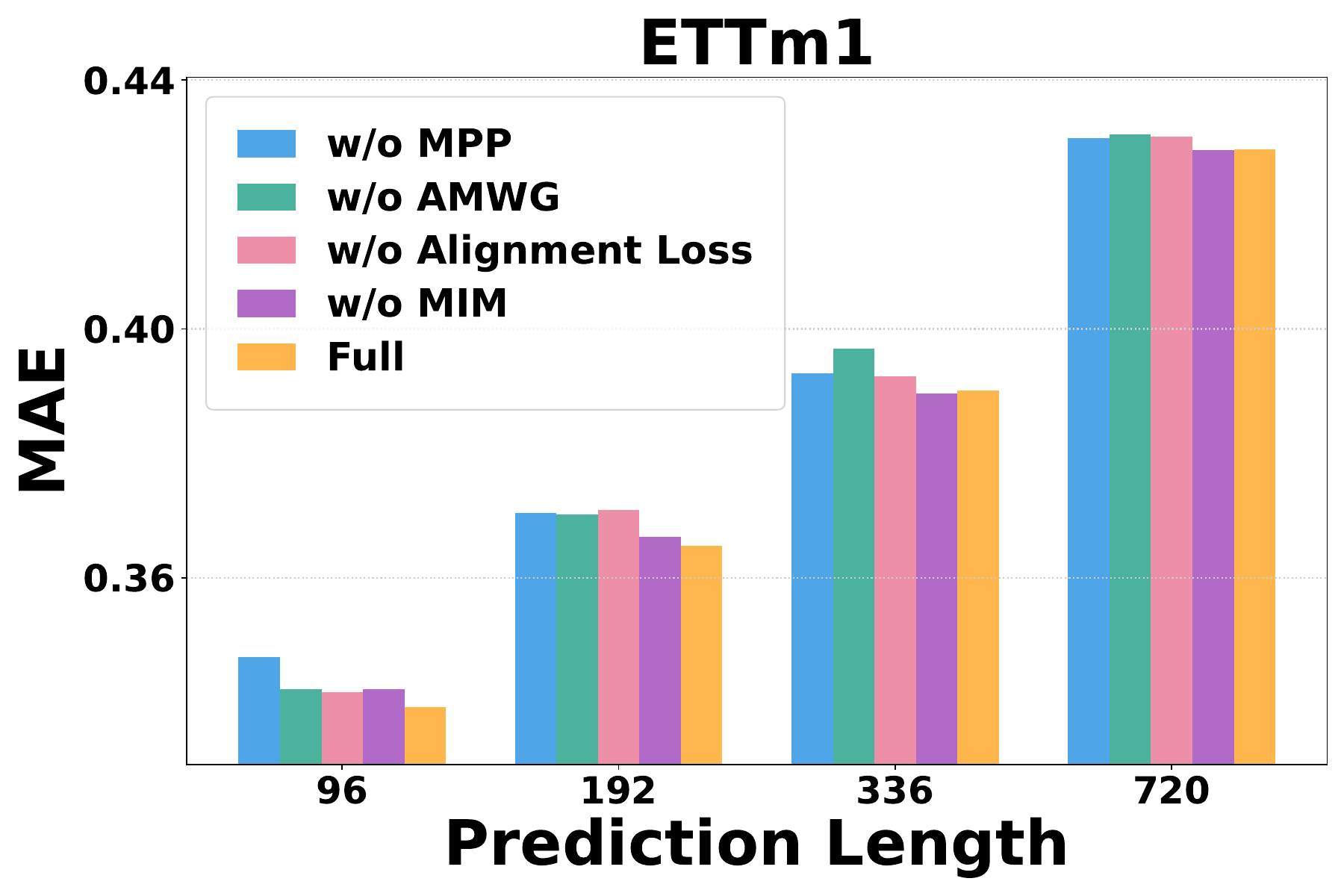}
        \label{fig:image3}
    \end{subfigure} \hfill
    \begin{subfigure}[b]{0.49\columnwidth}
        \centering
        \includegraphics[width=\columnwidth]{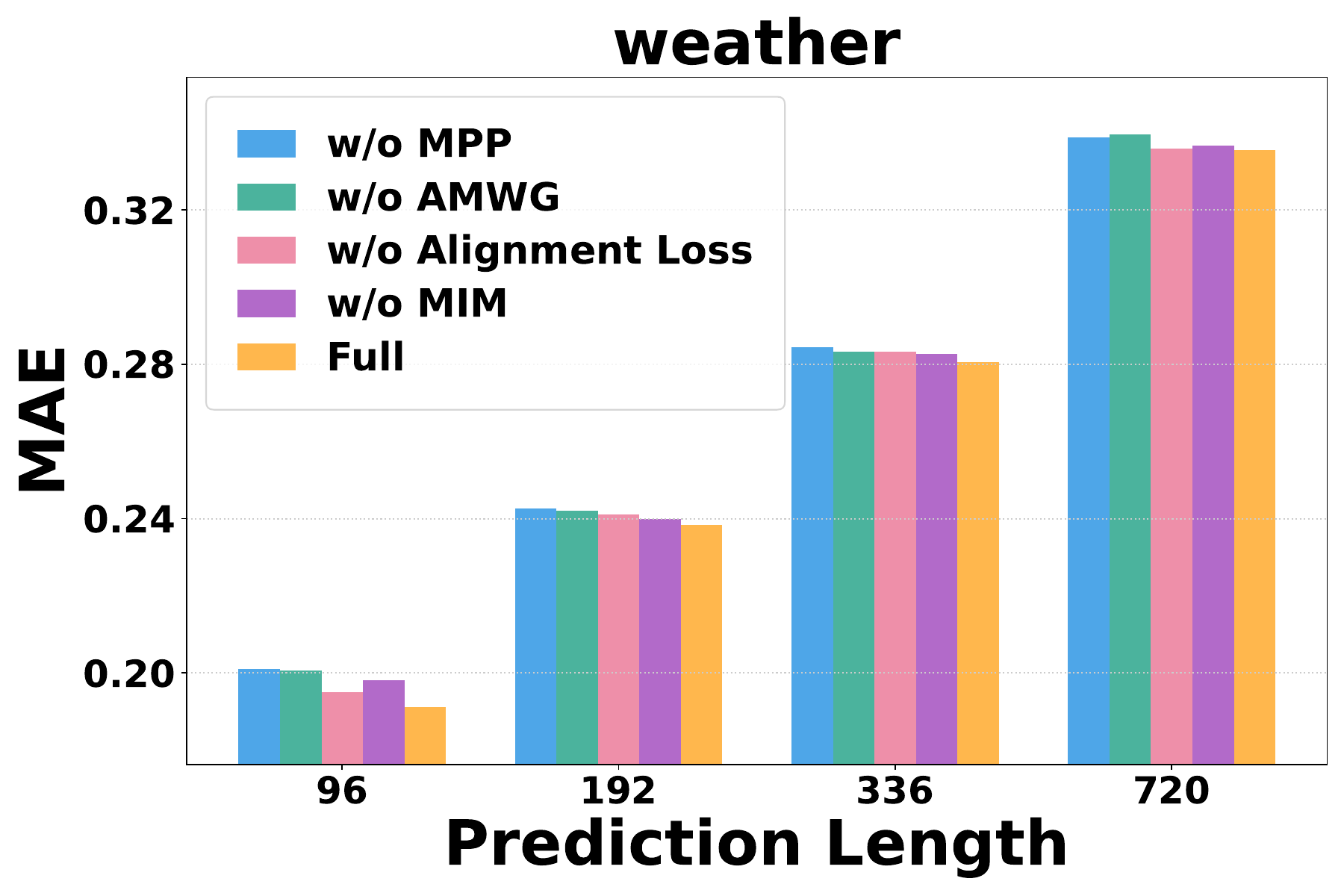}
        \label{fig:image4}
    \end{subfigure}

    \vspace{-2em} 

    \caption{Ablation of MPP, MIM, AMWG, and alignment loss on ETTm1 and Weather datasets.}
    \label{fig:abaltion}
    \vspace{-1.5em}
\end{figure}
MDMixer consists of three key modules: MPP, MIM, and AMWG, with the proposed alignment loss. To evaluate the effects of these components and methods on the model, we conduct ablation studies on the ETTm1 and Weather datasets. The corresponding results are presented in Figure~\ref{fig:abaltion}. "w/o MPP" indicates that MPP is not applied, meaning that only a single prediction head is used. In this configuration, since there is only one stream of prediction, MIM and AMWG are consequently not applicable and are also removed. "w/o MIM" indicates that MIM is not applied. "w/o AMWG" indicates that AMWG is not applied, and instead, predictions of different granularities are simply summed. "w/o Alignment Loss" indicates that the proposed alignment loss is not applied, and only MAE is used as the main loss. From the experimental results, it can be observed that removing any module from MDMixer leads to a significant performance degradation, which demonstrates the effectiveness of each module. A significant performance decline is observed when MPP is removed, underscoring the importance of explicitly disentangling features across multiple granularities. Likewise, eliminating AMWG leads to substantial degradation in model performance, highlighting the critical role of adaptive, channel-aware fusion over simple aggregation, as different variables depend on distinct levels of granularity. Also, the performance degradation caused by removing the alignment loss indicates that relying solely on the loss of the final prediction is insufficient to promote effective learning of intermediate representations.

\subsection{Hyperparameter Sensitivity}

\subsubsection{The influence of the number of heads and alignment loss weight}
\begin{figure}[h]
    \centering
    \begin{subfigure}[b]{0.49\columnwidth}
        \centering
        \includegraphics[width=\columnwidth]{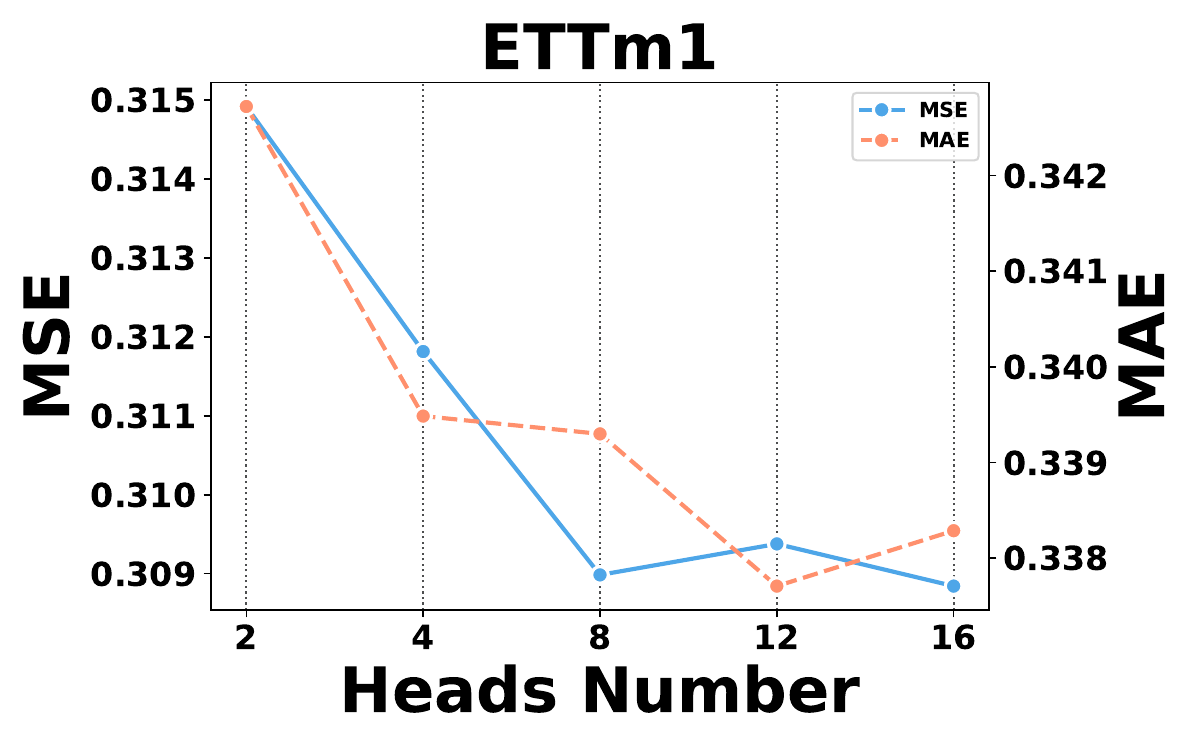}
        \label{fig:image1}
    \end{subfigure} \hfill
    \begin{subfigure}[b]{0.49\columnwidth}
        \centering
        \includegraphics[width=\columnwidth]{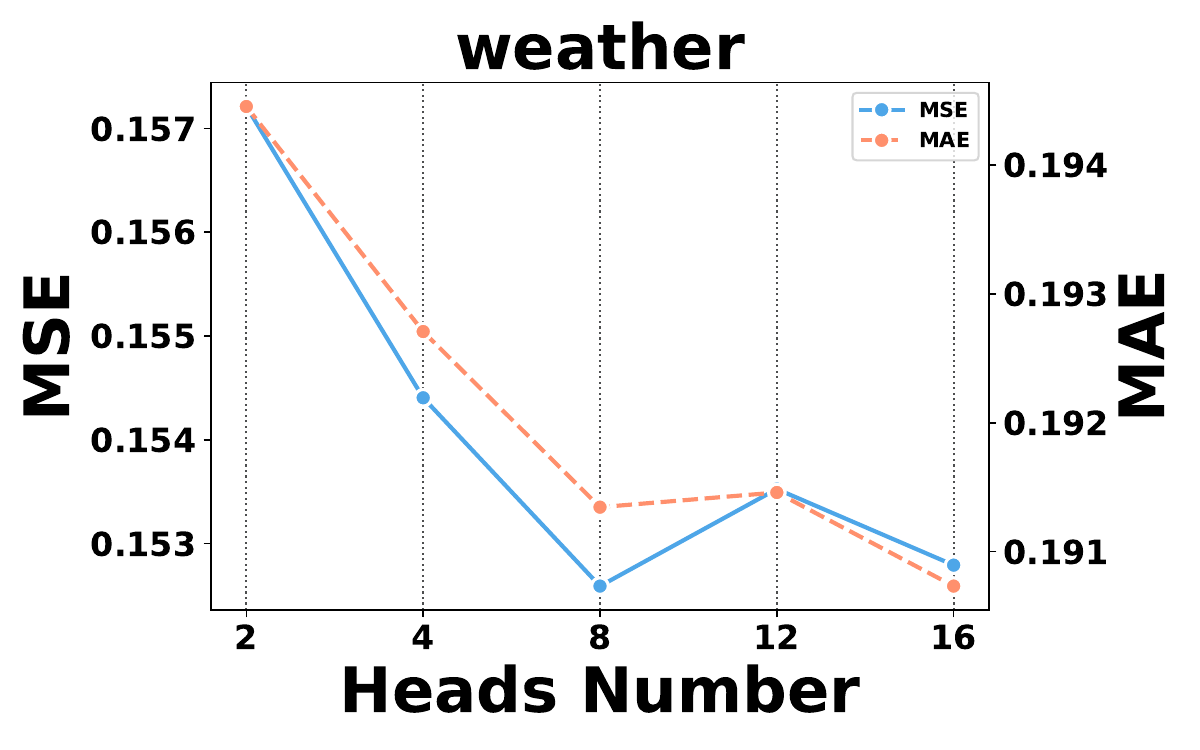}
        \label{fig:image2}
    \end{subfigure} \hfill
    \vspace{-2em}
    \caption{MSE scores with varying heads number $H \in \{2,4,8,12,16\}$.}
    \vspace{-1.5em}
    \label{heads number}
\end{figure}

\begin{figure}[h]
    \centering
    \begin{subfigure}[b]{0.49\columnwidth}
        \centering
        \includegraphics[width=\columnwidth]{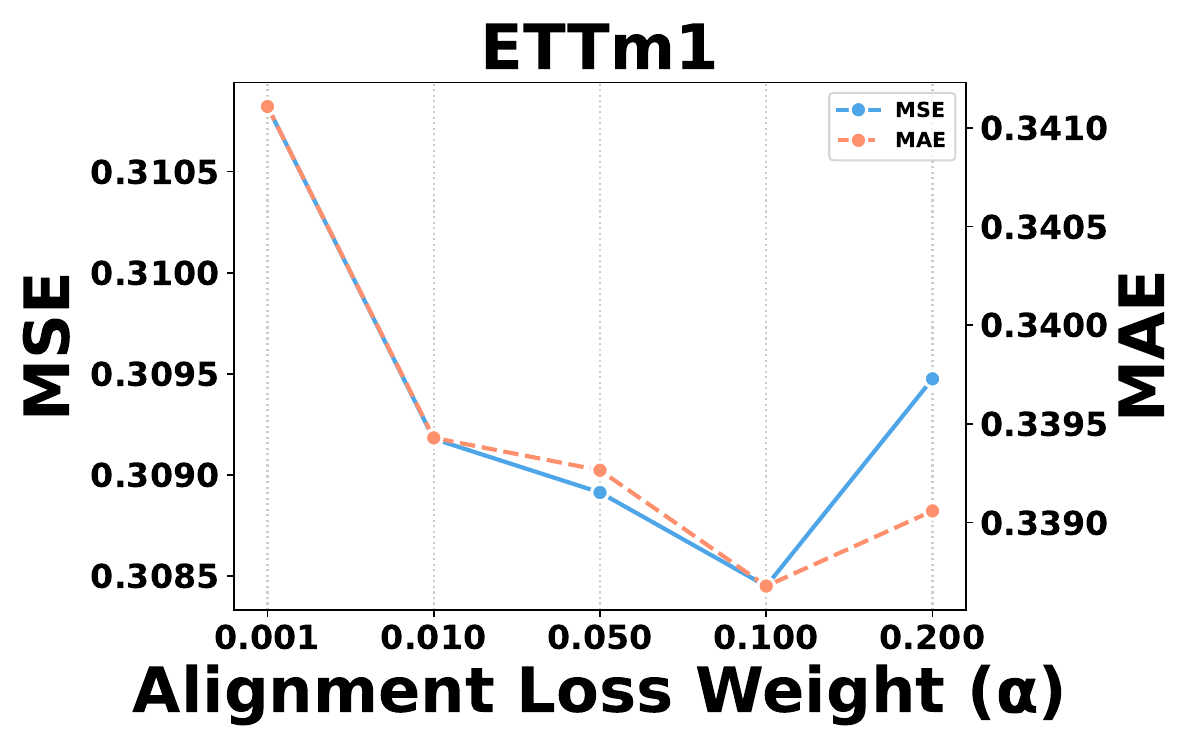}
        \label{fig:image1}
    \end{subfigure} \hfill
    \begin{subfigure}[b]{0.49\columnwidth}
        \centering
        \includegraphics[width=\columnwidth]{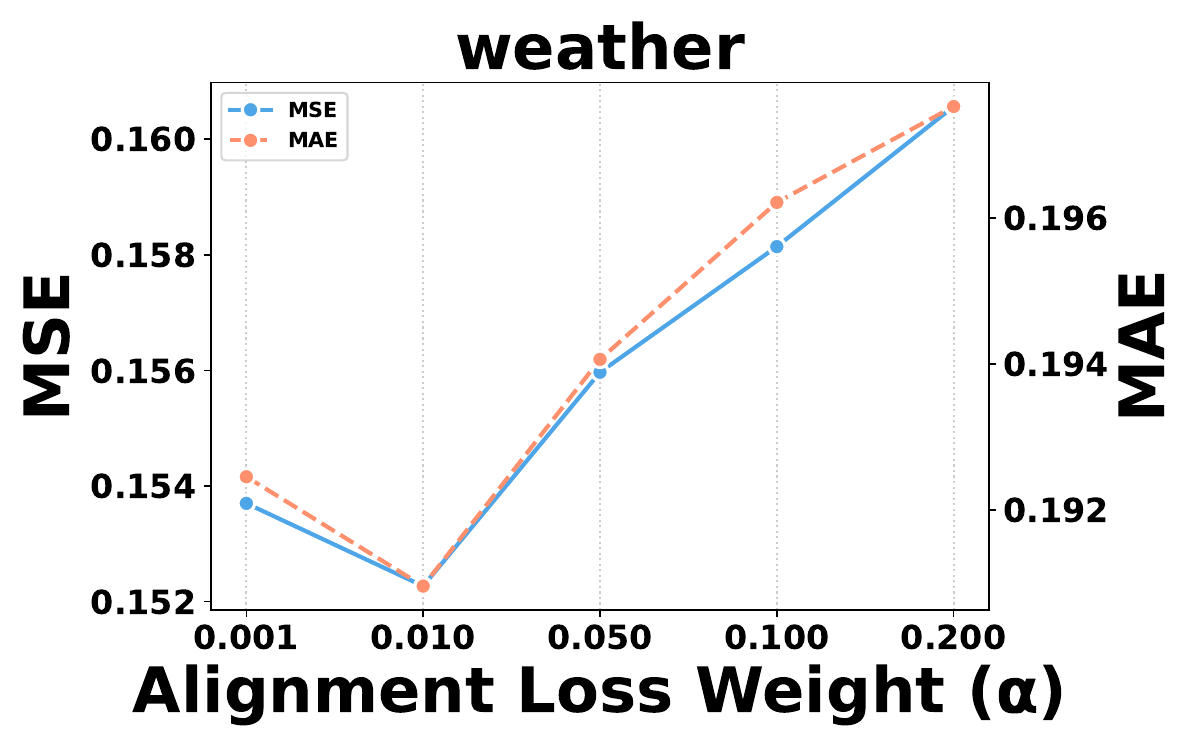}
        \label{fig:image2}
    \end{subfigure} \hfill
    \vspace{-2em}
    \caption{MSE scores with varying alignment loss weight $\alpha \in \{0.001,0.01,0.05,0.1,0.2\}$.}
    \vspace{-1.5em}
    \label{alpha}
\end{figure}

We conduct a sensitivity analysis on two important hyperparameters in our model: the number of prediction heads \( H \) and the alignment loss weight \( \alpha \), using the ETTm1 and weather datasets. We set both the input length and the prediction length at 96, varying $H$ from 2 to 16 and $\alpha$ from 0.001 to 0.2. Figure~\ref{heads number} illustrates the effect of heads number. It is observed that initially, increasing the number of heads significantly enhances performance, as the finer granularity enables more effective multi-scale feature extraction. When $H=8$, the performance becomes stable, indicating that MDMixer effectively captures diverse temporal patterns. Adding more heads beyond this point may lead to performance degradation due to overfitting. Consequently, we recommend using 8 or 12 heads to achieve an optimal balance between performance and computational efficiency. The results of $\alpha$ are shown in Figure~\ref{alpha}. When $\alpha$ is set too small (e.g., $\alpha=0.001$), the model produces suboptimal results, indicating insufficient intermediate-granularity guidance. As $\alpha$ increases, both MSE and MAE progressively decrease, reaching their minimum around 0.01 to 0.1, which suggests that moderate multi-granularity supervision effectively enhances the final prediction. However, an excessively large $\alpha$ (e.g., $\alpha=0.2$) results in slight performance degradation, likely because the model overemphasizes auxiliary alignment at the expense of final accuracy. In summary, we recommend setting $\alpha$ within the range of 0.01 to 0.1, depending on specific circumstances.

\begin{figure}[h]
    \centering
    \begin{subfigure}[b]{0.49\columnwidth}
        \centering
        \includegraphics[width=\columnwidth]{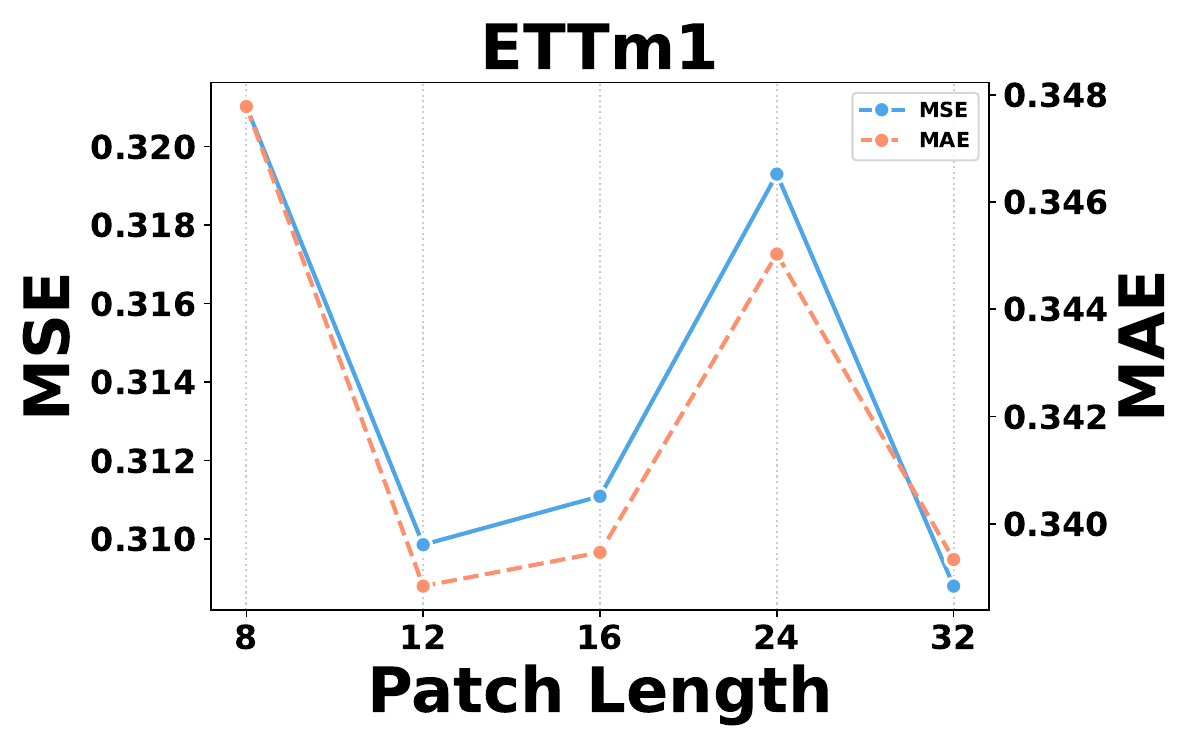}
        \label{fig:image1}
    \end{subfigure} \hfill
    \begin{subfigure}[b]{0.49\columnwidth}
        \centering
        \includegraphics[width=\columnwidth]{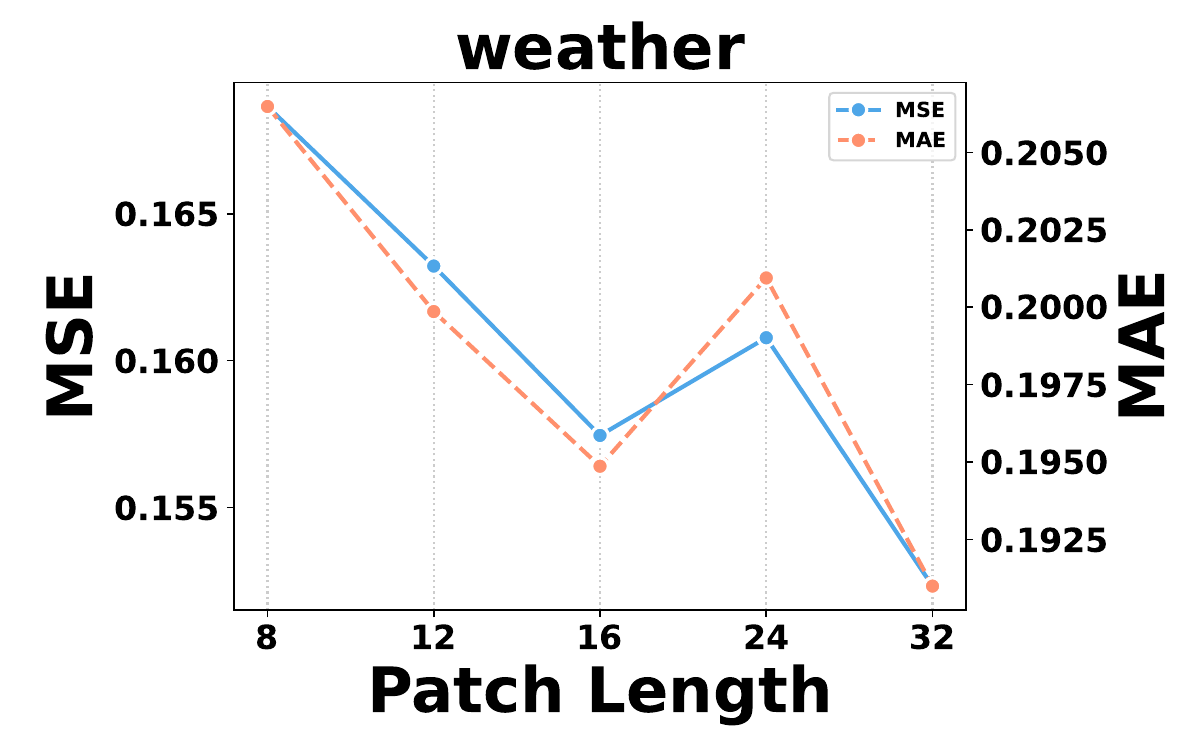}
        \label{fig:image2}
    \end{subfigure} \hfill
    \vspace{-2em}
    \caption{MSE scores with varying patch length $P \in \{8,12,16,24,32\}$.}
    \vspace{-1em}
    \label{patch len}
\end{figure}
\begin{figure}[]
\centering
    \begin{subfigure}[b]{0.49\columnwidth}
        \centering
        \includegraphics[width=\columnwidth]{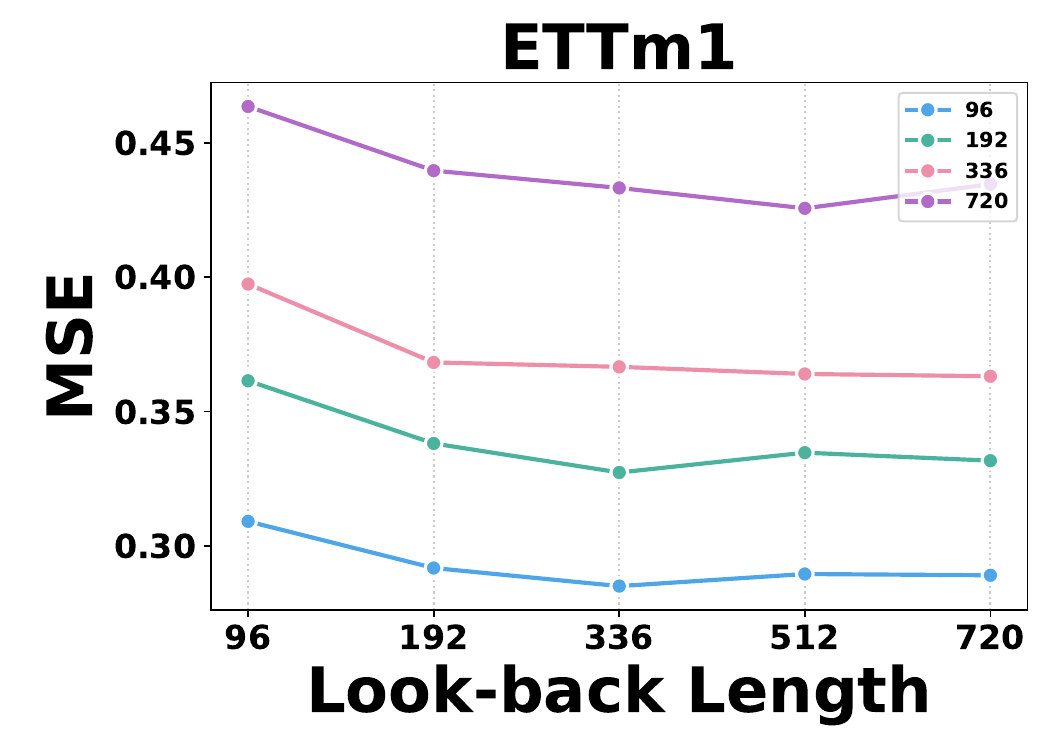}
        \label{fig:image1}
    \end{subfigure} \hfill
    \begin{subfigure}[b]{0.49\columnwidth}
        \centering
        \includegraphics[width=\columnwidth]{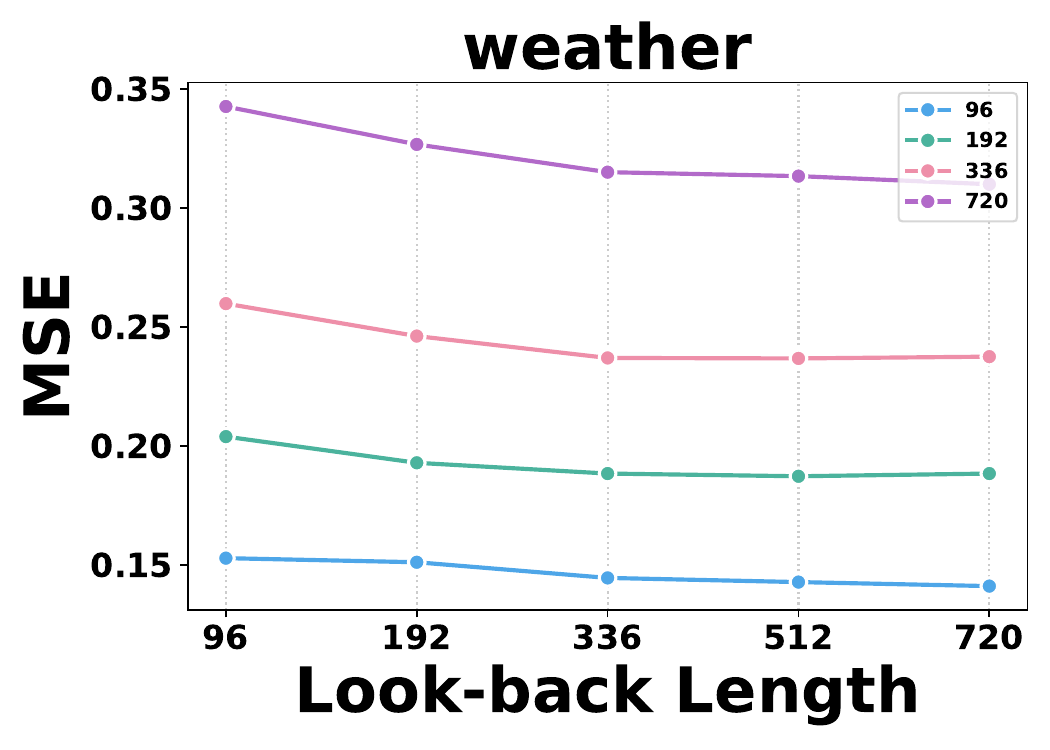}
        \label{fig:image2}
    \end{subfigure} \hfill
    \vspace{-2em}
\caption{The forecast error of MDMixer with varying lookback length $T \in \{96,192,336,512,720\}$. Each line represents a different fixed forecast horizon $F \in \{96,192,336,720\}$. }
\label{lookback length}
\vspace{-1em}
\end{figure}
\subsubsection{The influence of patch length}
To investigate the effect of patch length $P$ on the model, we varied $P$ from 8 to 32 using the ETTm1 and Weather datasets. We set the stride length to half of the patch length, and both the input and output sequence lengths were fixed at 96, and the experimental results are presented in Figure~\ref{patch len}. The results show that as $P$ increases, the model’s prediction error initially reduces, reaching a relatively low value around 16, and achieving the lowest MSE at 32. It is important to note that excessively small patch lengths increase the number of model parameters. Therefore, considering both model performance and computational efficiency, we select $P=32$.

\subsubsection{The influence of look-back length}
In LTSF tasks, longer look-back lengths encapsulate more information regarding trend and cyclical patterns, and a powerful model should be able to leverage this information for more accurate predictions. However, previous studies have shown that the prediction performance of transformer models does not necessarily improve as the input length increases, which can be attributed to the distracted attention on the growing input~\cite{PatchTST, DLinear,iTransformer}. To assess whether our model can benefit from longer input sequences, we evaluate the performance of MDMixer under varying look-back lengths in Figure~\ref{lookback length}. The results demonstrate that as the look-back length increases, the prediction error of MDMixer consistently decreases, suggesting that MDMixer is able to capture richer information from longer historical observations to make more accurate predictions. Additionally, it is observed that MDMixer maintains strong performance even with shorter look-back lengths, highlighting the robustness of our model when faced with fewer observational data.

\subsection{Model Efficiency Analysis}
\begin{figure}[]
\centering
    \begin{subfigure}[b]{0.49\columnwidth}
        \centering
        \includegraphics[width=\columnwidth]{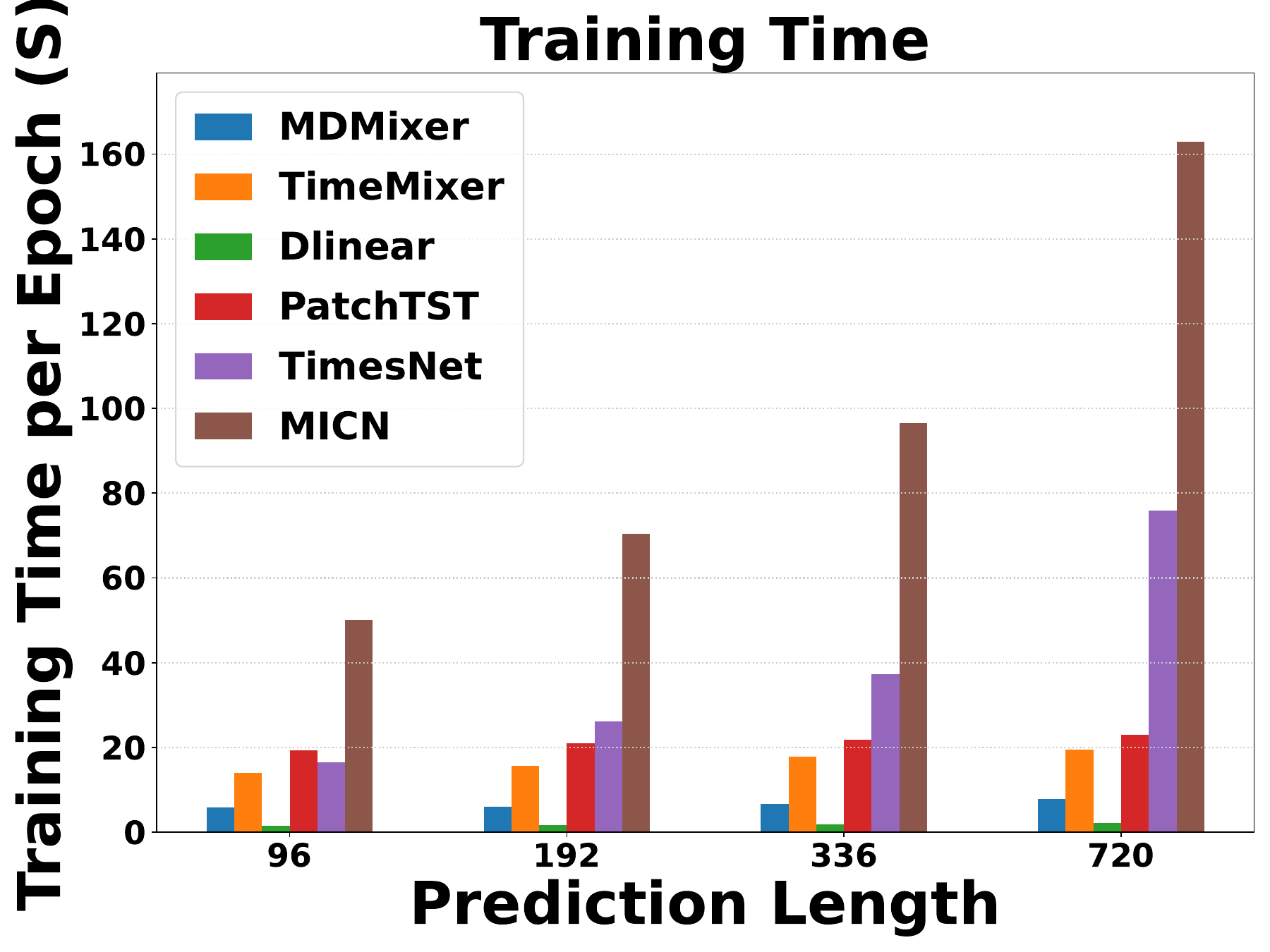}
        \label{fig:image1}
    \end{subfigure} \hfill
    \begin{subfigure}[b]{0.49\columnwidth}
        \centering
        \includegraphics[width=\columnwidth]{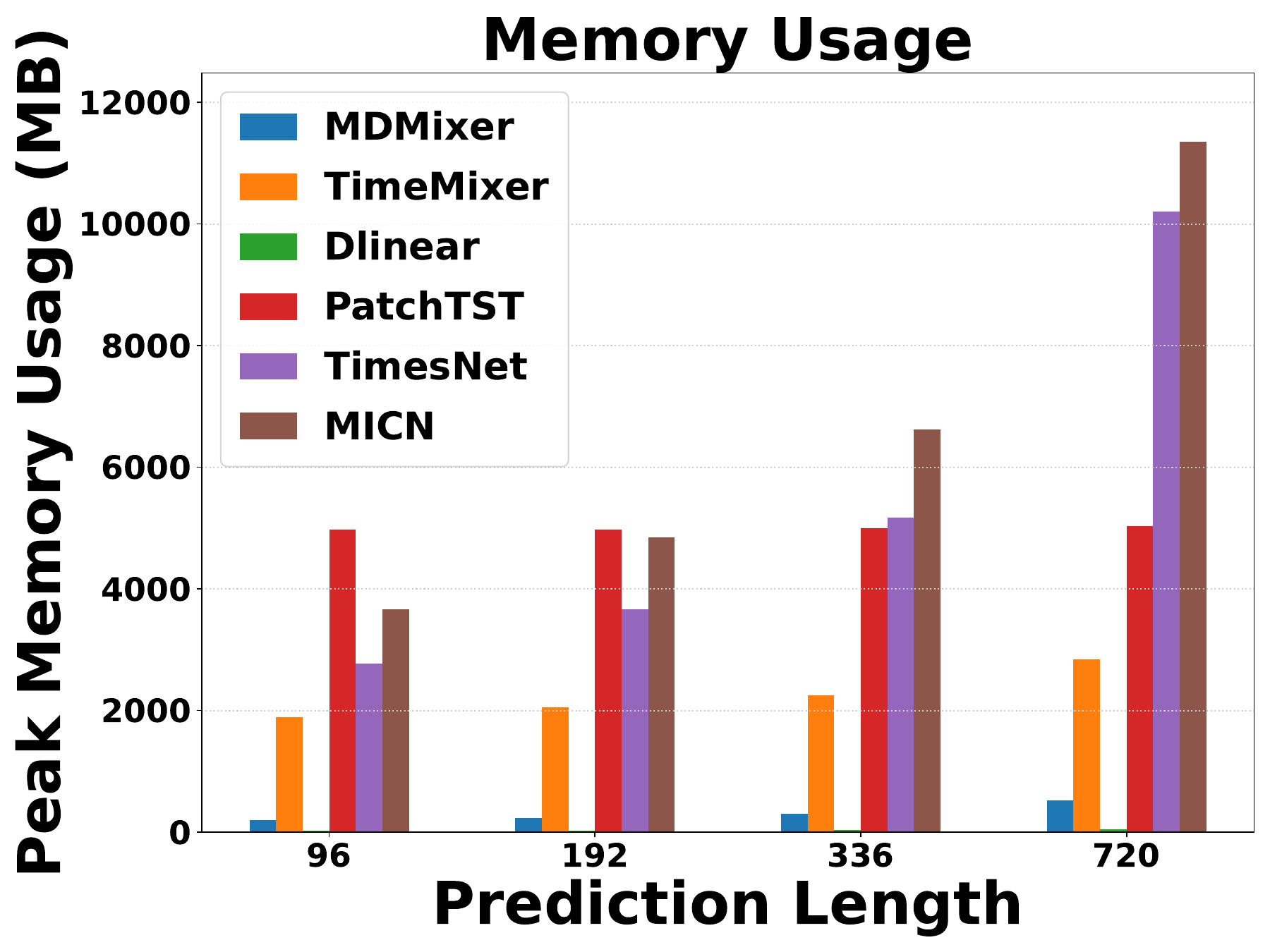}
        \label{fig:image2}
    \end{subfigure} \hfill
    \vspace{-2em}
\caption{The comparison of training time and memory usage between MDMixer and other models.}
\label{efficiency}
\end{figure}
MDMixer is a lightweight model. Figure~\ref{efficiency} presents experimental results comparing our model against several recent state-of-the-art models in terms of training time per epoch and peak GPU memory usage on the ETTm1 dataset. For a fair comparison, all models are configured with the same batch size and look-back length and evaluated on the same hardware. The results demonstrate that, aside from DLinear, our model achieves the lowest training time and memory usage among the models tested. Compared to TimeMixer, MDMixer achieves average reductions of approximately 61\% in training time and 86\% in memory usage. Benefiting from its lightweight, MLP-based architectural design, MDMixer exhibits substantial improvements in computational efficiency over Transformer-based models. In particular, compared to PatchTST, MDMixer also reduces training time and memory usage by approximately 69\% and 94\%, respectively. These experimental results highlight that MDMixer successfully balances prediction accuracy with computational efficiency.

\begin{figure}[h]
    \centering
    \begin{subfigure}[b]{0.8\columnwidth}
        \centering
        \includegraphics[width=\linewidth]{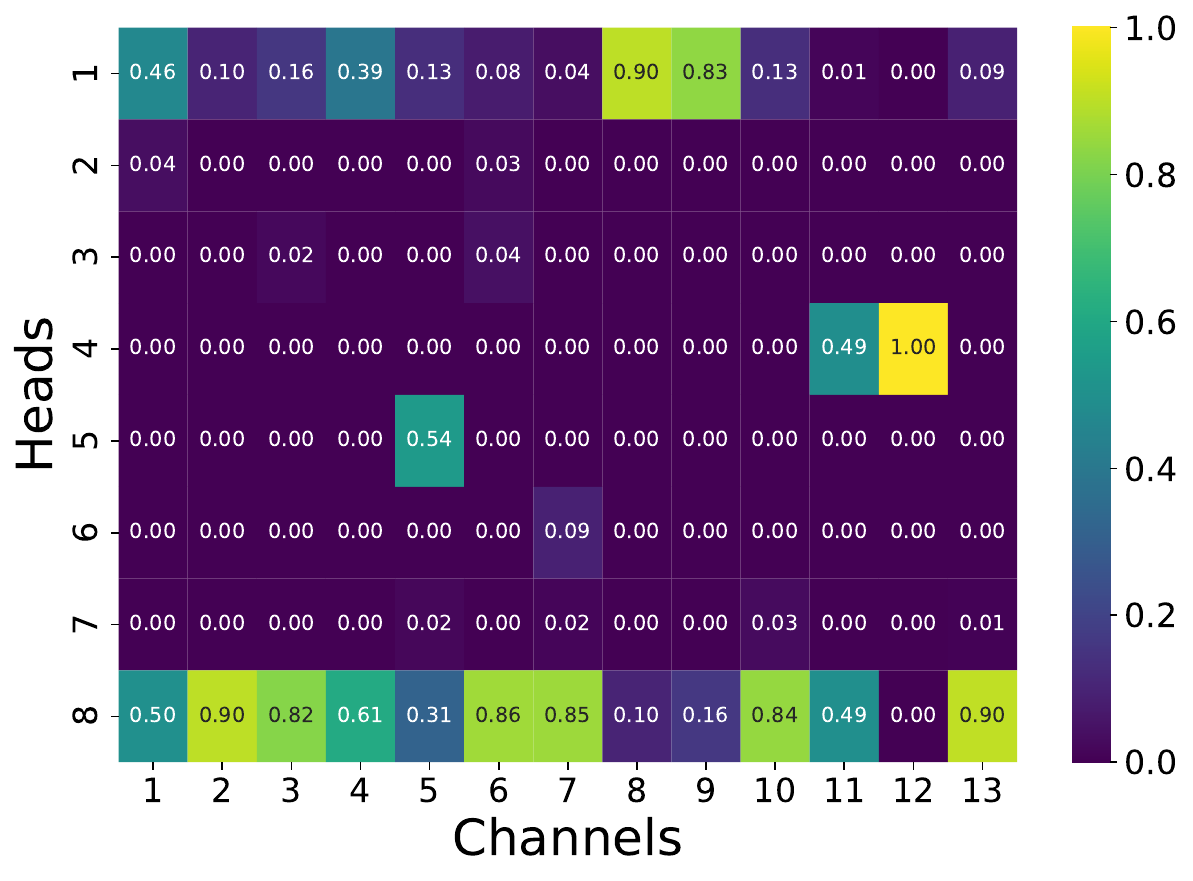}
        \label{fig:image1}
    \end{subfigure}
    
    \vspace{-1em}

    \caption{Visualization of the weights assigned by AMWG to different granularities across various channels on the electricity dataset.}
    \label{heatmap}
    \vspace{-1em}
\end{figure}

\subsection{Model Interpretability Analysis}
To gain a deeper understanding of how MDMixer dynamically integrates information of different granularities, we plot the weight heatmaps generated by AMWG on several channels of the electricity dataset. As shown in Figure~\ref{heatmap}, higher head indices correspond to finer granularity. From the figure, we can observe that the gating mechanism exhibits sparsity, meaning that the contributions of different granularities to the model’s prediction are not equally important; the model selectively focuses on information from specific granularities. In the case of $H=8$, significant weights are often observed on Head 1 (coarsest) and Head 8 (finest) across various channels, indicating that MDMixer tends to utilize information from both ends of the granularity spectrum for prediction. This result supports our initial observation (Figure~\ref{mixing}), namely that coarse-grained predictions capture overall seasonal and trend patterns, while fine-grained predictions extract short-term fluctuations. Notably, the emphasis on specific granularities varies substantially across different channels. This suggests that our model can selectively extract the most relevant granularity information for each channel, effectively down-weighting contributions from less informative granularities. These outcomes underscore the importance of adaptive and channel-aware fusion, explaining why simple fixed-weight fusion methods, such as direct addition (as shown in ablation studies), can lead to performance degradation. In summary, AMWG enables MDMixer to adaptively integrate multi-granularity information, enhancing both interpretability and predictive performance in LTSF tasks.
\section{Conclusion and Future Work}
In this paper, we propose MDMixer, which is designed to address the intricate entanglement of multi-granularity information and temporal patterns in time series. Our novel Multi-granularity Parallel Predictor and Mixer, as well as Adaptive Multi-granularity Weighting Gate, disentangle and integrate distinct temporal patterns. The dual-branch architecture we designed enhances the predictive performance of the model for various components of the time series. Extensive experiments on multiple real-world benchmarks validate MDMixer's state-of-the-art performance, superior computational efficiency, and enhanced model interpretability. In the future, we will further refine dynamic multi-granularity fusion mechanisms and investigate more explicit methods for information interaction between channels to enhance the model's ability in long-term forecasting.

\bibliographystyle{ACM-Reference-Format}
\bibliography{arxiv}

\end{document}